\documentclass{article}
\usepackage{natbib} 
\usepackage[table]{xcolor}
\usepackage{longtable}
\usepackage{url}
\usepackage{pdflscape}
\usepackage{hyperref}
\usepackage{authblk}
\usepackage{tikz}
\usepackage[draft]{todonotes}
\usepackage{booktabs}
\usepackage{subcaption}
\usepackage{datetime}
\usepackage{caption}
\usepackage{amsmath}

\usepackage[symbol]{footmisc}

\providecommand{\keywords}[1]
{
  \small	
  \textbf{\textit{Keywords---}} #1
}

\date{}
\title{Inferred vs traditional personality assessment: are we predicting the same thing?}
\author[1]{Pavel Novikov}
\author[1,2]{Larisa Mararitsa\footnote{Corresponding author.
E-mail address: larisa.mararitsa@humanteq.io}}

\author[1]{Victor Nozdrachev}
\affil[1]{Humanteq, Moscow, Russian Federation}
\affil[2]{National Research University Higher School of Economics, Saint-Petersburg, Russian Federation}

\def\checkmark{\tikz\fill[scale=0.4](0,.35) -- (.25,0) -- (1,.7) -- (.25,.15) -- cycle;}

\begin{document}

\newcommand{\hgrayline}{\arrayrulecolor{lightgray}\hline\arrayrulecolor{black}}

\maketitle

Machine learning methods are widely used by researchers to predict psychological characteristics from digital records. To answer whether automatic personality estimates retain the properties of the original traits, we reviewed 220 recent articles.

First, we put together the predictive quality estimates from a subset of the studies which declare separation of training, validation, and testing phases, which is critical for ensuring the correctness of quality estimates in machine learning(\cite{hastie2009elements, zheng2015evaluating, kumar2019machine}). Only 20\% of the reviewed papers met this criterion. To compare the reported quality estimates, we converted them to approximate Pearson correlations. The credible upper limits for correlations between predicted and self-reported personality traits vary in a range between 0.42 and 0.48, depending on the specific trait. The achieved values are substantially below the correlations between traits measured with distinct self-report questionnaires. This suggests that we cannot readily interpret personality predictions as estimates of the original traits or expect predicted personality traits to reproduce known relationships with life outcomes regularly.

Next, we complement quality estimates evaluation with evidence on psychometric properties of predicted traits. The few existing results suggest that predicted traits are less stable with time and have lower effective dimensionality than self-reported personality. The predictive text-based models perform substantially worse outside their training domains but stay above a random baseline. The evidence on the relationships between predicted traits and external variables is mixed. Predictive features are difficult to use for validation, due to the lack of prior hypotheses. Thus, predicted personality traits fail to retain important properties of the original characteristics. This calls for the cautious use and targeted validation of the predictive models.


\section{Introduction}

 The idea that it is possible to identify psychological characteristics of a person from digital traces of behaviour dates back to 2005 (\cite{argamon2005lexical}), and accumulated evidence has been growing with the amount of available data. There are currently hundreds of studies that infer psychological characteristics using digital records (see Sections \ref{section:literature}, \ref{section:surveys}). Despite this, it is still difficult to summarize progress in the field and state the success of the modern solutions. In this paper, we approach this question by assembling a broad collection of recent studies on the automatic inference of psychological characteristics and review the various aspects of assessing prediction validity, limiting our attention to a subset of the research concerned with predicting stable personality characteristics. 

The paper is organized as follows. In Section \ref{section:personality}, we give a summary of psychological research on personality, highlighting some aspects that may be of relevance for automatic prediction research. In Section \ref{section:literature}, we describe the details of the process of aggregating relevant papers. In section \ref{section:surveys}, we put the present study into the context of existing literature reviews on automatic personality prediction. In Section \ref{section:data}, we review datasets that are used in automatic personality inference research. In Section \ref{section:validity_framework}, we look at the diverse practices applied by the researchers to investigate the quality of trained personality prediction algorithms and conceptualize the practices using a framework of psychometric instrument validation. In Section \ref{section:performance}, we specifically address classification and regression performance estimates. In Section \ref{section:conclusions}, we conclude with final observations and discussion.

\section{Personality}
\label{section:personality}

\subsection{Five factor model (Big-5)}

The Five-factor Model is a framework of personality measurement based on a long research tradition, which is widely accepted by psychologists (\cite{john2008paradigm, matz2016models}). The key premise of the model is that stable personal characteristics are continuously distributed in a five-dimensional space. This model has emerged from factor analyses of thousands of personal attributes that are embedded in natural languages (\cite{john2008paradigm}). 

These personality dimensions (traits) are measured using one of many available instruments, or questionnaires (some examples are \cite{costa1992revised,john2008paradigm, perugini2002analyzing, goldberg1999broad, rammstedt2007measuring, markey2009brief, gosling2003very,konstabel2017measuring, soto2017next}). A personality assessment instrument must meet several established criteria (\cite{furr2017psychometrics}. See also \cite{soto2017next} for a recent example of an instrument development). The measurement is performed by collecting `degree-of-agreement' answers to a set of questions. The answers are then converted into numeric estimates through a simple sum. Five traits are usually called extraversion (or surgency), neuroticism (or emotional stability), openness to experience (or intellect and imagination), agreeableness, and conscientiousness (\cite{soto2017next, ipip_constructs}). 

The names differ between instruments, as well as their exact semantics. The correlation estimates of the same traits measured with different Big-5 instruments are usually in the range of 0.6--0.9 \cite{john2008paradigm, soto2017next, ipip_comparison}, although the estimates are substantially lower if instruments that come from other traditions are considered \cite{pace2010similar}.

There are many well-supported findings that give an outline of the advantages and limitations of the five-factor model. The same factor structure is reproduced across countries and languages \cite{kajonius2017personality}. The Big-5 dimensions have been shown to relate to a multitude of life outcomes \cite{soto2019replicable}. Test-retest reliability (a correlation between the same trait estimates using the same questionnaire with an interval of several weeks between measurements) is estimated to lie between 0.674 and 0.851, depending on the specific trait and the instrument \cite{Gnambs2014AMO}. Rank order stability (order correlation of the same respondent group) over a period of several years is estimated to be 0.6, or even more if the adolescent-adulthood transition is not included (\cite{costa2019personality}). At the same time, traits have been shown to gradually evolve with age \cite{costa2019personality} and relate to gender \cite{schmitt2008can, Giolla2018SexDI}. There are varying estimates of trait intercorrelations, but it is common for estimated trait-pair correlations to exceed 0.2, and a value of 0.3 is occasionally reached (\cite{park2020meta}).

Big-5 estimates can be obtained through self-reported measures as well as questionnaires completed by others (apparent personality). Self-reported measures are correlated with measures of highly acquainted raters. Strangers show much less agreement with self-reports while agreeing with each other (\cite{Connelly2010AnOP}).

Summing up, we can assume with a high degree of confidence that the Big-5 traits capture some general human characteristics. These characteristics are relatively stable with time and are related to important life outcomes.
At the same time, it is important to remember that these test measurements are inherently noisy (as illustrated by test-retest reliability estimates). Traits are not completely independent from each other or from demographic characteristics,  and they can vary based on the instrument employed. Also, when considering labeling by external raters, we must be aware that the problem we pose, albeit valid, is not the same as self-reported personality prediction.

\subsection{Myers-Briggs Type Indicator}

Myers-Briggs Type Indicator (MBTI) \cite{king2020myers} is another popular personality assessment tool. According to The Myers-Briggs Company website \cite{mbti_instrument}, it is used by more than 88 percent of Fortune 500 companies in 115 countries. Its popularity among social network users allows to mine relatively big datasets with MBTI annotations shared by users \cite{Verhoeven2016TwiStyAM, Bassignana2020PersonalITYAN}.

MBTI assessment divides respondents into mutually exclusive groups for each of four ``preference pairs": Extraversion (\textbf{E}) -- Introversion (\textbf{I}), Sensing (\textbf{S}) -- Intuition (\textbf{N}), Thinking (\textbf{T}) -- Feeling (\textbf{F}), and Judging (\textbf{J}) -- Perceiving (\textbf{P}). MBTI instruments are commercial tools that are restricted for purchase to people with a relevant degree or those who had attended a certification program.

MBTI has not been studied as extensively as the Big-5, yet there is some limited research supporting its validity (\cite{randall2017validity,capraro2002myers}). Also, the MBTI factors have been noticed to overlap with the Big-5 factors (\cite{furnham2003relationship}).

There are criticisms in the academic community concerning MBTI (\cite{stein2019evaluating, king2020myers}). Several of these criticisms are addressed on the official site of the Myers-Briggs Company (\cite{mbti_criticism_response}).

\subsection{Boundaries of personality research}

There are other personality models and instruments that have been proposed over the years and show varying overlap with the Big-5.  For example, there are Eysenck personality questionnaire (\cite{aziz2001comparison}) and Cattell's 16PF questionnaire (\cite{Gerbing1991The1R, rossier2004hierarchical,aluja200516pf5}) that preceded the emergence of the Big-5. 

There are also more recent models. One of them is the Dark Triad model, integrating offensive yet non-pathological personality traits: narcissism,  Machiavellianism,  and psychopathy. Another example is the HEXACO (\cite{howard2020discriminant}) model, which extends the Big-5 with an additional honesty/humility trait.

Moreover, the exact definitions of personality differ among researchers \cite{piechurska2020personality}. For example, there are arguments whether or not the concept of personality should include intelligence (\cite{deyoung2011intelligence}) and basic values (\cite{parks2015personality}). For the present work, we stick with the  broad interpretation and consider all the stable psychological traits relevant for the study. 

We exclude from the consideration works devoted to the prediction of the changeable psychological states, such as mood and emotions, as well as prediction of mental disorders (\cite{Dzedzickis2020HumanER, Shatte2019MachineLI,rohani2018correlations}).

\section{Literature aggregation} 
\label{section:literature}

Our focus was on collecting a sample of studies devoted to automatic personality inference from digital data which is as broad and as exhaustive as possible rather than unbiased. Thus, we adopted a method based on two procedures that heavily rely on human decisions: \textit{search} and \textit{expansion}. 

\textit{Search} refers to querying the open academic search engines, \textit{google scholar} (\cite{google_scholar})  and \textit{semanticscholar} (\cite{fricke2018semantic}) with key phrases like ``personality prediction", ``psychoinformatics", and ``computational psychology". Whenever we encountered new phrases that could potentially refer to automatic personality prediction problems, we also added them to the search. 

\textit{Expansion} refers to searching an item from our sample on Semantic Scholar and looking for relevant titles in ``Citations", ``References", and ``Related Papers" sections. We prioritized candidates for expansion based on the year of the publication (newest first) and the category of the paper (see below). 

While collecting the sample, we were manually categorizing the papers we had found. Papers introducing datasets, as well as literature surveys were assigned the highest priority for expansion. Personality prediction studies were marked for further analysis. 

 We continued the procedure until our search saturated as most of the expansion iterations started to give no new candidates. As a result, we collected more than 800 candidate papers. 
 We limited our focus to the papers devoted to automatic trait inference from digital records that were published in or after 2017. We excluded non-English papers and papers in which both the data and the labels come from self-reports. 
 
This gave us a sample of 218 papers focused on building the prediction algorithm and two more papers that examined pre-trained models. \autoref{table:bigrams} shows the most popular bigrams in the titles of the selected papers.

For the most part, the present study is based on the entire sample. Additional screening was performed in the Section \ref{subsection:estimates}.

\begin{table}
\centering
\begin{tabular}{|p{5cm}|p{1cm}|}

\hline
Bigram & Count \\
\hline
\hline
personality traits & 10 \\
\hline
personality trait & 8 \\
\hline
automatic personality & 8 \\
\hline
big five & 6 \\
\hline
five personality & 6 \\
\hline
predicting personality & 6 \\
\hline
neural networks & 6 \\
\hline
personality prediction & 6 \\
\hline
machine learning & 4 \\
\hline
personality classification & 4 \\
\hline

\end{tabular}
\caption{Frequent bigrams in the titles from the sample of 220 papers investigating  automatic personality prediction (without stop-words)}
\label{table:bigrams}
\end{table}

\section{Related reviews}
\label{section:surveys}

There is a considerable number  of studies that aim to summarize various aspects of personality inference research. \cite{Agastya2019ASL} and \cite{Mehta2019RecentTI} have surveyed personality prediction studies based on several data sources and modalities. Both have paid special attention to deep learning-based approaches. \cite{Jacques2018FirstIA} and \cite{escalera2018guest} have limited the scope of the review to apparent personality prediction.

Substantial attention is paid to personality prediction based on social media. \cite{OngExploringPP}, \cite{Bhavya2018PersonalityIF}, and \cite{Kaushal2018EmergingTI}  have made systematic summaries of the literature with a focus on distinct parts of the prediction pipeline. \cite{Tay2020PsychometricAV} pay special attention to the validity of automatic personality estimates from the perspective of personality assessment. In a series of meta-analyses, \cite{Azucar2018PredictingTB}, \cite{settanni2018predicting} and \cite{marengo2020digital}, look at some studies on personality prediction from the social network in order to compare their performance as well as identify study characteristics that affect it. 

Some studies take a broader look. \cite{Burr2019CanMR} focused on surveying the studies on prediction of psychometric properties including affect, autism, psychopathy, and more. \cite{Pan2018AutomaticallyIH} have extended the scope of prediction targets of analysis to include ethnicity, occupation, interests, human values, etc.  \cite{Pan2018AutomaticallyIH} have made a synthesis of diverse kinds of research related to personality manifestation in technology use. 

Most of the works above compile performance estimates  with a mix of the performance metrics used in the original studies.
The variety of measures does not allow them to be sorted out to identify the state of the art. \cite{Azucar2018PredictingTB}, \cite{settanni2018predicting} and \cite{marengo2020digital} alleviate this problem by converting some of the metrics into approximate Pearson correlations. They also investigate the moderating effects of several study characteristics on performance. 

Looking at their findings, however, one can notice several surprising things. In the meta-analyses performed by \cite{settanni2018predicting} and \cite{marengo2020digital}, several older papers are well ahead of the newer ones in performance. Even more surprisingly, \cite{marengo2020digital} have found that many studies (6 out of a sample of 23) do not employ a separate test set for evaluation, yet there is no significant effect of the test set evaluation on performance. Considering this, we decided to build on the idea of bringing the performance to the single scale while paying special attention to the correctness of the evaluation procedure. 

 To further broaden the understanding of the prediction quality, we have also followed \cite{Tay2020PsychometricAV} by looking at diverse aspects of the psychometric validity of inferred personality in addition to predictive performance. We applied these ideas to analyse a subset of the personality inference literature which is several times larger than the ones in the previous studies. 
At the same time, while it is common for review papers to include an analysis of the learning algorithms, it is out of the scope of the present study. This is because our study focuses on the achieved performance and validity themselves, rather than on the identification of the features that influence them.

\section{Datasets}
\label{section:data}

Much work on personality inference is based on private data. According to our sample, more than 40\% of studies involve data collection, and it can easily be an underestimation (since expansion by citation favours papers which are interlinked with the datasets in the citation graph).
However, there are many works that rely on existing datasets. 
Here, we focus on reusable datasets. We consider datasets to be reusable if they appear in two or more papers or if their creators explicitly mention that they may be shared (possibly with a limited group of researchers). The datasets we have identified can be roughly divided into several categories, although some fall into more than one. The categories are listed below. For more information regarding the datasets, see \autoref{table:datasets}.

\subsection{Essays}
The stream-of-consciousness essay dataset is one of the first ones utilized in personality prediction research (\cite{Mairesse2007UsingLC, argamon2005lexical}) and is still very much in use (\cite{MehtaBottomUpAT, Salminen2020EnrichingSM, Zhang2020DrawingOT}). 
This dataset consists of about 2500 stream-of-consciousness essays written by students and annotated with Big-5 scores.  
Researchers who use this dataset usually cite \cite{pennebaker1999linguistic}. However, a brief inspection of the paper and the dataset suggests a discrepancy in the time of collection. The data collection process in the paper took place between 1992 and 1998, while the data suggests the period between 1997 and 2004. We would rather argue that the stream-of-consciousness dataset is an extension of a dataset introduced by \cite {argamon2005lexical}. In any case, both papers report collecting at least two essays per student. This fact can potentially endanger the validity of a test set evaluation. 
Several other personality-annotated essay datasets also exist: in Dutch (\cite{Luyckx2008PersonaeAC, verhoeven2014clips}), Russian (\cite{Litvinova2018RusNeuroPsychOC}), and Spanish (\cite{RamrezdelaRosa2018TxPIuAR} --- handwritten essays with transcriptions).

\subsection{Social media with self-reports}
The largest dataset that is used in personality prediction research is the one collected by the myPersonality project (\cite{my_personality, Kosinski2013PrivateTA}). This dataset contains data of Facebook users who filled in a personality questionnaire and/or some other questionnaires within the myPersonality Facebook App, and who also opted to donate their Facebook data to the research. myPersonality data used to be provided to other scholars with anonymized data and could be used for non-commercial academic research, yet this practice ended in 2018. 
It is hard to estimate the size of the dataset exactly since it is not static and different research focused on different subsets. Still, it is safe to say that the total number of respondents is in the millions (\cite{Asadzadeh2017AnalyzingFA,Xue2018DeepLP, zhang2018situation}) and the size of the annotated dataset is in tens of thousands. For example, \cite{Xue2018DeepLP} report that this dataset contains 115,864 users, whose default language is English and for whom the Big Five personality trait scores are available. 
A small subset of myPersonality dataset with Facebook status updates and social network graph properties of 250 users annotated with the Big-5 traits was released in the Workshop on Computational Personality Recognition: Shared Task (\cite{celli2013workshop}). This is the most reused dataset in the papers we have selected (almost 30 entries). 

There are several more personality-annotated datasets of posts in several social networks: Twitter (\cite{rangel2015overview, Salem2019PersonalityTF}), Facebook (\cite{Ramos2018BuildingAC, Hall2017AmIW}), and Sina microblog (\cite{Liu2016DeepLF}). 
The PsychoFlickr dataset (\cite{cristani2013unveiling}), in contrast, is built of liked images (200 per person) from Flickr. It is annotated with self-reported traits, as well as attributed by unacquainted judges.   

\subsection{Social media with mined labels}
Recently, the attention of researchers has turned towards personality data that is shared by users in social media, which led to the creation of mined personality-annotated datasets from Twitter (\cite{Verhoeven2016TwiStyAM, Plank2015PersonalityTO}), Reddit (\cite{Gjurkovic2020PANDORATP, gjurkovic2018reddit}), Youtube (\cite{Bassignana2020PersonalITYAN}), and the PersonalityCafe forum (\cite{kaggle_mbti}). Among them, there is the largest personality-annotated dataset currently open for research (TwiSty --- \cite{Verhoeven2016TwiStyAM}) with 18,168 entries. Unlike most of the other datasets, almost all of the datasets in this category provide only MBTI types, illustrating the vast disparity in personality measure popularity between the research community and general public. The only exception is the Reddit-based PANDORA dataset (\cite{Gjurkovic2020PANDORATP}).

\subsection{Apparent personality}
There are several datasets with personality annotations made by unacquainted judges. By far, the largest is the first impressions dataset (\cite{ponce2016chalearn}). It comprises 10,000 clips (with an average duration of 15s) extracted from more than 3,000 different YouTube high-definition (HD) videos of people facing a camera and speaking in English. The clips have been annotated by mTurkers (\cite{mturk}), who ranked pairs of videos along the Big-5 dimensions. Cardinal scores have been obtained by fitting a Bradley-Terry-Luce model (\cite{hunter2004mm}). 
This dataset has been released in ECCV ChaLearn LAP 2016 challenge. Importantly, it has been divided into train (60\%), development (20\%), and test (20\%), and this division is respected in later studies. 
An extended version of the dataset (\cite{escalante2018explaining}) with an additional variable indicating whether the subject should be invited to a job interview or not have also been used in ChaLearn Workshop on Explainable Computer Vision Multimedia and Job Candidate Screening Competition @CVPR17 and @IJCNN17 (\cite{chalearn17_competition}). 

Notice that, conceptually, the personality here is considered a property of a clip rather than a person. Clips are annotated individually, and more than 80\% of validation and test clips have counterparts from the same video in the train set. The same approach is shared by other video (\cite{celiktutan2014maptraits}) and speech (\cite{Zhang2017SocialPE, Mohammadi2012AutomaticPP}) clip datasets.

In contrast, in the dataset introduced by \cite{biel2013youtube}, each example is a 1-minute video of a single vlogger. There is also a transcribed text version of this dataset (\cite{biel2013hi}). This dataset was used in the workshop on computational personality recognition 2014. 

There are also datasets that combine both apparent and self-reported personality annotations (\cite{cristani2013unveiling, Mehl2006PersonalityII}).

\subsection{Phone data}

Several data collection initiatives (\cite{wang2014studentlife,
meng2014analyzing, aharony2011social, stachl2020predicting}) focused on various types of phone logs. This data type can include WiFi and GPS-based location, phone use, Bluetooth-based contact estimation, and other modalities. 

\subsection{Laboratory experiments}
There is also a group of datasets collected using some special-purpose equipment (wearable sensors, high-quality cameras, sound-proof booths, etc.), with relatively large amounts of high-quality annotated data for a small number of subjects (the largest includes 139 subject pairs) (\cite{correa2018amigos, levitan2015cross, celiktutan2017multimodal, alameda2015salsa, Mehl2006PersonalityII, Subramanian2018ASCERTAINEA,sanchez2013emergent, dotti2018behavior, Khan2020VyaktitvAM, koutsombogera2018modeling, Butt2020MultimodalPT}).

\section{Validity of inferred personality}
\label{section:validity_framework}

The standard way to evaluate a machine learning algorithm is to measure the agreement between the prediction and the target variables. However, in the context of automatic prediction, the trained algorithm can be viewed as an alternative personality assessment instrument, and as such, it should arguably conform to the same validity requirements (\cite{furr2017psychometrics}). In connection with this, \cite{Bleidorn2019UsingML} proposed a validation framework that includes several distinct components. 

\textit{Reliability} refers to the consistency of estimates across time, raters, or content. \textit{Generalizability} is the degree to which the scale accurately measures the construct it is designed to measure in novel contexts. In the scope of machine learning, it directly translates to cross-domain generalization (\cite{domain_generalization}). \textit{Factorial validity} implies that the same factor structure of a multidimensional measure is reproduced across contexts. \textit{Criterion validity} refers to the degree to which a measure predicts relevant outcomes. \textit{Convergent validity}  is the degree to which the measures of the same construct correlate with each other. It corresponds directly to predictive performance. \textit{Discriminant validity} refers to the degree to which the correlation of a measure with different measures does not exceed expected limits. \textit{Incremental validity} refers to the degree to which a measure predicts a relevant outcome beyond what is known based on other measures. \textit{Content validity} implies that the test indicators should be justified based on the underlying theory. In \cite{Bleidorn2019UsingML}, the authors admit that content validity is the hardest to translate to machine learning problems because of their data-driven nature. The authors see a possible solution in making additional efforts to interpret the models and to distinguish features consistent with the definition of traits from the surprising ones.

With this framework in mind, we looked through the collected studies and compiled approaches that were used to evaluate personality prediction. Our search showed that most studies from our sample only estimated convergent validity (predictive performance). It is addressed in detail in Section \ref{section:performance}. We found no techniques that could be categorized as measuring incremental validity and found only one study that approached discriminant validity \cite{RoblesGranda2020JointlyPJ} by including a correlation table between predictions and a range of self-reported measurements. The other findings are listed below.

\subsection{Content validity}

Although it is difficult to require content validity as theoretical justification from machine learning algorithms, its simplified form --- interpretability --- is probably the most studied. Here we list some of the studies that address interpretability issues. 

\cite{Stachl2020PredictingPF} explored the importance of phone activity features on predicting Big-5 personality by measuring a performance drop after variable permutations. \cite{MehtaBottomUpAT} built word clouds with the size of words based on SHAP-values.  \cite{Rssola2019PersonalityRI} explored word importance derived from capsule neural networks. \cite{Dotti2019BeingTC} investigated structural feature activations with a network trained to recognize personality from records of group behaviour. 
\cite{Park2018WhenSD} reported predictive features for inferring personality using HTTP(s) traffic. 
\cite{Segalin2017SocialPT} illustrated important visual features for personality prediction from favorite images.
\cite{Stanford2017PredictionOP, Gltrk2017VisualizingAP, Ventura2017InterpretingCM} presented several ways to show significant visual features for predicting apparent personality from vlog videos. 

In the cases above, there is no formal way to distinguish intuitive and surprising features. \cite{Lynn2020HierarchicalMF} approached this problem by asking two personality researchers to separate pieces of text that were highly indicative of user personality from neutral pieces and compared their decisions with those of the trained model. The experts showed an agreement with the model for openness to experience and agreeableness, while the results for conscientiousness, neuroticism, and extraversion were no better than chance.

\subsection{Factorial validity}

The stability of multidimensional factor structure is not directly addressed in any of the studies in our sample. Some studies report intercorrelations of self-reported traits (for example, \cite{Kachur2020AssessingTB, Xiong2019PrivacyFriendlyPR, Tareaf2019FacialBasedPP, Stachl2020PredictingPF, 
Anselmi2019GenuinePR}). It is much rarer to report prediction intercorrelations or structure of correlations between self-reported and predicted traits. 

There are a few exceptions. \cite{Gjurkovic2020PANDORATP} report correlations  between self-reported and predicted traits including cross-trait and cross-instrument correlations. \cite{RoblesGranda2020JointlyPJ} include correlations between all self-reported and predicted measures (the Big-5 scales and 12 more measures). \cite{Jayaratne2020PredictingPU} included predicted trait intercorrelations. 

\begin{figure}
\hfill
\begin{subfigure}[t]{.475\textwidth}
\includegraphics[width=\linewidth]{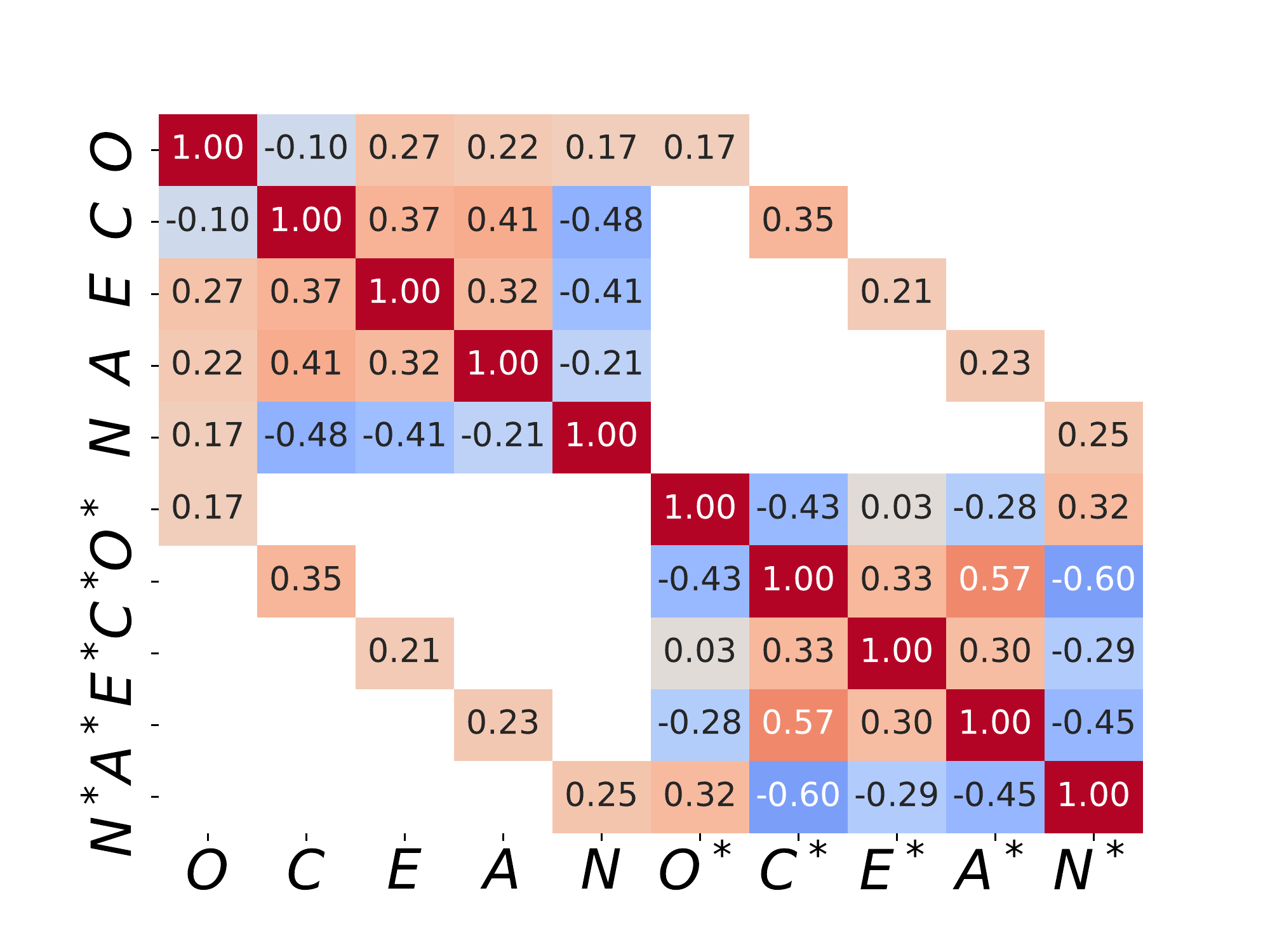}
\caption{}
\end{subfigure}
\hfill
\begin{subfigure}[t]{.475\textwidth}
\includegraphics[width=\linewidth]{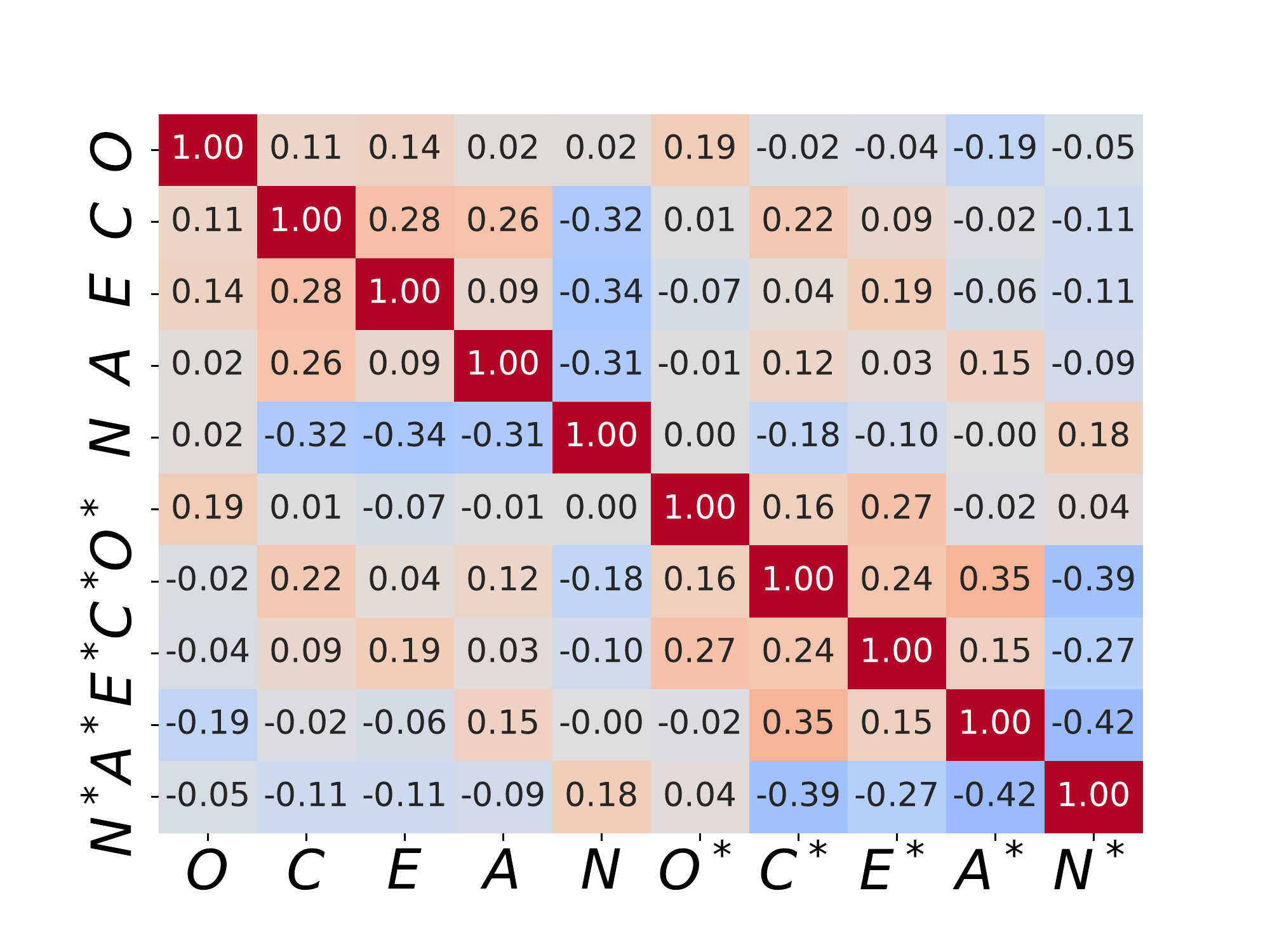}
\caption{}
\end{subfigure}
\hfill
\caption{Correlations of self-reported and predicted traits. Asterisk marks predicted traits. a) data from  \cite{Kachur2020AssessingTB} averaged for men and women. b) data from  \cite{Liu2016AnalyzingPT}}
\label{figure:intercorrelations}
\end{figure}

\cite{Kachur2020AssessingTB} and \cite{Liu2016AnalyzingPT} reported both self-reported and predicted trait intercorrelations along with some cross-correlations (see \autoref{figure:intercorrelations}) The pattern of self-reported trait intercorrelations is similar to the one in the predictions. At the same time, predictions are more intercorrelated.  Also, the scale of the predicted traits and their self-reported counterparts is often smaller than intercorrelations with other traits. These observations may indicate a shift in the semantics of the prediction relative to the original traits.

\cite{Calefato2019ALI} used an external automatic personality prediction tool to detect the Big-5 traits for a group of software developers. They performed principal component analysis of the five traits and found only two significant components that accounted for 72\% of the total variation.

\subsection{Generalizability}

\cite{Akrami2019AutomaticEO} explored cross-domain transferability of a model trained on online forum texts annotated by unacquainted psychology students to two text datasets annotated with self-reported personality. Thus, both text domain and annotation type varied. The authors found that performance for all predictions, as measured by $R^2$, was below zero. 

\cite{Carducci2018TwitPersonalityCP} trained the personality prediction model on Facebook texts and evaluated on an independent dataset from Twitter. Unfortunately, the results are reported in terms of mean squared error and it is hard to judge the eventual performance.

\cite{An2018LexicalAA} used two personality datasets, one based on Facebook texts and the other consisting of transcribed audio clips. They performed cross-assessment of model quality and found that performance dropped, relative to the best in-domain models (from the accuracy of 0.67 to 0.60 and from 0.61 to 0.53) and yet they are still above or equal to a baseline. 

A similar experiment was performed by \cite{Jaidka2018DifferencesIS} on stress and empathy (as traits, rather than states) prediction  from Facebook and Twitter. Their results were mostly in line with the other researchers, although they found that Twitter models generalize on Facebook better than the other way around.

\cite{Siddique2017BilingualWE} explored cross-lingual generalization by using bilingual word embedding trained on a separate parallel text corpus. In their experiments, the cross-lingual model outperformed the monolingual model for 3 of 5 scales. 

\subsection{Reliability and criterion validity}

Explicit examination of predicted personality trait correspondences with outcomes that are expected from theory (criterion validity) and prediction stability (reliability) are rarely performed in the studies. Thus, in this section, we included any procedures that explored how predicted traits relate to external variables and parameters.

\cite{Li2019QuantitativePP} is the only study in our sample that addressed conventional reliability and criterion validity. The authors built a model to predict personality from EEG records. They recorded the data again for a subset of participants 19--78 days later to calculate test-retest reliability. For another subset of users, they collected some additional self-reported data and calculated correlation with both reported and predicted personality traits. In both validation procedures, the scores for all predicted traits were significant, albeit lower than those for self-reported personality.

\cite{Calefato2019ALI} showed that the median predicted trait variance with time was not significantly different from zero, although this does not say anything about individual variability. Additionally, the authors found evidence that developers who exhibited higher levels of predicted openness and agreeableness were more likely to become contributors to Apache projects.

\cite{kern2019social} applied an external automatic personality prediction tool to a set of Twitter profiles grouped by occupation and showed that occupations produce distinct trait patterns. For example, in their experiments, top GitHub contributors tended to have high openness to experience and low agreeableness, while chief
information officers were characterized with high openness to experience and emotional stability. The authors further built classifiers to distinguish between 10 professions based on the personality profile and achieved accuracy above 0.7.

\cite{Gladstone2019CanPT} built a model to predict personality from spending records and evaluated the moderating effect of sociodemographic variables on predictive performance. The authors found no effect from age, salary, or total spending, although participants living in areas that are highly deprived turned out to be difficult to predict.

Similarly, \cite{Gilpin2018PerceptionOS} compared the performance of apparent personality prediction from audio on distinct sociodemographic groups. They found differences between male and female populations, as well as journalists and non-journalists. 

\cite{Zhang2017PhysiognomyPT} trained a model to predict personality (measured with Cattel 16PF instrument) from face images. They further collected images corresponding to two occupations --- teachers and entertainers --- that were not used in training and calculated the predicted traits. Entertainers got high scores on reasoning  and liveliness, and low on emotional stability, while teachers got high scores in liveliness and extroversion. 

\cite{Titov2019FullscalePP} trained models for predicting personality from social media profiles and used them to profile subscribers of official groups of two car brands. They found significant differences between the audiences in most traits.

\section{Classification and regression performance}
\label{section:performance}

All the papers in our sample that focus on model construction include some estimates of conventional prediction quality, also referred to as convergent validity in the validation framework in Section \ref{section:validity_framework}. Different studies adopt different formal problem definitions and different evaluation metrics. In this section, we attempt to summarize this variety. In Section \ref{subsection:metric_overview}, we review the most popular metrics for personality prediction evaluation and highlight some difficulties in interpreting them. In Section \ref{subsection:metric_conversion}, we describe a method used to convert between the metrics. In Section \ref{subsection:estimates} we describe an additional screening procedure and compile the estimates.

\subsection{Labels and performance metrics}
\label{subsection:metric_overview}

There is an overwhelming homogeneity in the research in considering personality prediction as several separate trait prediction tasks. We have only been able to identify five studies from our sample that deviated from this trend:
\cite{Fralenko2019CorrelationAA} includes a prediction of a Big-5 trait with the largest score. \cite{Lai2020AutomaticPI} divides the sample into four categories based on two of the Big-5 traits. Finally, \cite{Gjurkovic2018RedditAG}, \cite{Cui2017SurveyAO}, and  \cite{Buraya2018MultiviewPP} address the prediction of an exact MBTI type.

However, the specifics of prediction task formulation vary drastically. Some studies approach trait prediction as a binary classification using median (\cite{Farnadi2020InferringAG}), mean (\cite{Yuan2020PersonalityEA}), or other (\cite{Klados2020AutomaticRO}) threshold for score discretization. Some adopt 3-class (low, medium, high) formulation (\cite{Zhou2018ExtrovertsTD, Bhin2019RecognitionOP}). Others exclude the middle class and consider binary classification limited to extreme groups (\cite{Segalin2017WhatYF, Wei2017BeyondTW}). Finally, regression formulation is also often adopted (\cite{Nurek2020HawkesmodeledTP, Bhin2019RecognitionOP}). 

The choices of metrics for evaluating performance also vary. \autoref{table:metrics} shows how often most popular metrics were reported within our sample. This suggests that, in many cases, metrics from two papers would not allow direct comparison. In addition, there are some issues specific to particular metrics that complicate interpretation. 

\begin{table*}
\centering
\begin{tabular}{|l|l|l|}

	\hline
	Metric & Total & Single metric\\
    \hline
  	Accuracy & 96 & 40\\
  	\hline
 	F1 & 63 & 18\\
 	\hline
 	Precision & 35 & 1\\
 	\hline
 	Recall & 35 & 1\\
 	\hline
 	ROC AUC & 14 & 5\\
 	\hline
 	(R)MSE & 43 & 15\\
 	\hline
 	MAE & 22 & 5\\
 	\hline
 	Correlation & 32 & 8\\
 	\hline
 	$R^2$ & 17 & 1\\
 	\hline
 	1-MAE & 14 & 11\\
 	\hline

\end{tabular}
\caption{Usage frequency of popular performance evaluation metrics in the sample of 218 studies devoted to training automatic personality prediction models. We show separately a number of papers that report a single metric.}
\label{table:metrics}
\end{table*}

One difficulty shared by the classification approaches, especially binary, is that the baseline quality depends on the proportion of the classes. Self-reported trait distributions are usually bell-shaped (\cite{Leonardi2020MultilingualTP,Anselmi2019GenuinePR,Khan2020VyaktitvAM, Stankevich2019PredictingPT, Darliansyah2019SENTIPEDEAS, Li2020CRNetAD}). They are also discrete because of the nature of the measurement instruments. Consequently, any split near the middle of the distribution will inevitably make one class bigger by a non-negligible amount and shift the baselines away from 0.5. This implies, for example, that without the specifics, we cannot be sure whether the precision of 0.3 is trivial or that accuracy of 0.7 is not.

Some caution is necessary when interpreting results reported in terms of accuracy. This word is sometimes used as a synonym for performance and can consequently refer to some metrics other than the proportion of correct class attributions. Notably, the metric introduced in the ECCV ChaLearn LAP 2016 challenge (\cite{ponce2016chalearn}) along with the first impressions dataset was called ``mean accuracy". This metric is defined by the organizers as mean relative absolute error subtracted from one, although studies based on the dataset often define it simply as 1-MAE (\cite{Wei2018DeepBR, Kaya2018MultimodalPT, Principi2019OnTE}). Sometimes the confusion between the classification accuracy and first impressions' mean accuracy finds its way into reviews (\cite{Mehta2019RecentTI}).

The interpretation of results reported in terms of F1, precision, or recall is complicated by the fact that their complete specification requires that it either explicitly chooses the positive class or that it fixes an averaging procedure (macro or micro averaging). This specification is often omitted. 

MAE and RMSE are quite popular regression metrics. Despite this, there are some problems with using them.  \cite{Sumner2012PredictingDT} have noted that estimates may be over-optimistic since most true labels tend to be close to the mean. \cite{Azucar2018PredictingTB} had to exclude papers that reported MAE or RMSE from meta-analysis since this information was not sufficient to estimate correlations. It is also important to note that the interpretation of MAE and RMSE is very much dependent on the scale of the data. 

It is also worth noting that although in most cases correlation stands for Pearson correlation, some studies report Spearman rank correlation (\cite{Segalin2017ThePW,Wei2017HowSD}) or disattenuated Pearson correlation (\cite{Lynn2020HierarchicalMF, Gladstone2019CanPT}). Some studies do not give the specifics.

\subsection{Performance metric conversion}
\label{subsection:metric_conversion}

To make the performance estimates comparable, \cite{Azucar2018PredictingTB} employed several conversion procedures to transform them into Pearson correlation.
Each of the procedures relies on a set of simplifying assumptions about target and prediction variables, like joint Gaussian or logistic distributions.

However, extending this approach to new metrics or measurement scenarios requires new formulas, possibly with their own assumptions. Instead, we decided to use a unified approach based on the Monte Carlo simulation. First, we assumed joint Gaussian distribution for target and prediction (such as self-reported and predicted extraversion). We further assumed the distribution to have a zero mean and the covariance matrix given by 
$$\Sigma = 
\begin{bmatrix} 
1 & r \\
r & 1
\end{bmatrix}
$$ Here $r$ is a parameter that corresponds to the correlation between target and prediction. For each value of $r$ between -1 and 1 with a step of 0.01, we performed a random simulation generating 10 million 2-dimensional points, and we used this sample to calculate a set of agreement metrics.  When calculating binary classification metrics we always assumed the same discretization threshold for label and prediction variables. 

\begin{figure}
\hfill
\begin{subfigure}[t]{.475\textwidth}
\includegraphics[width=\linewidth]{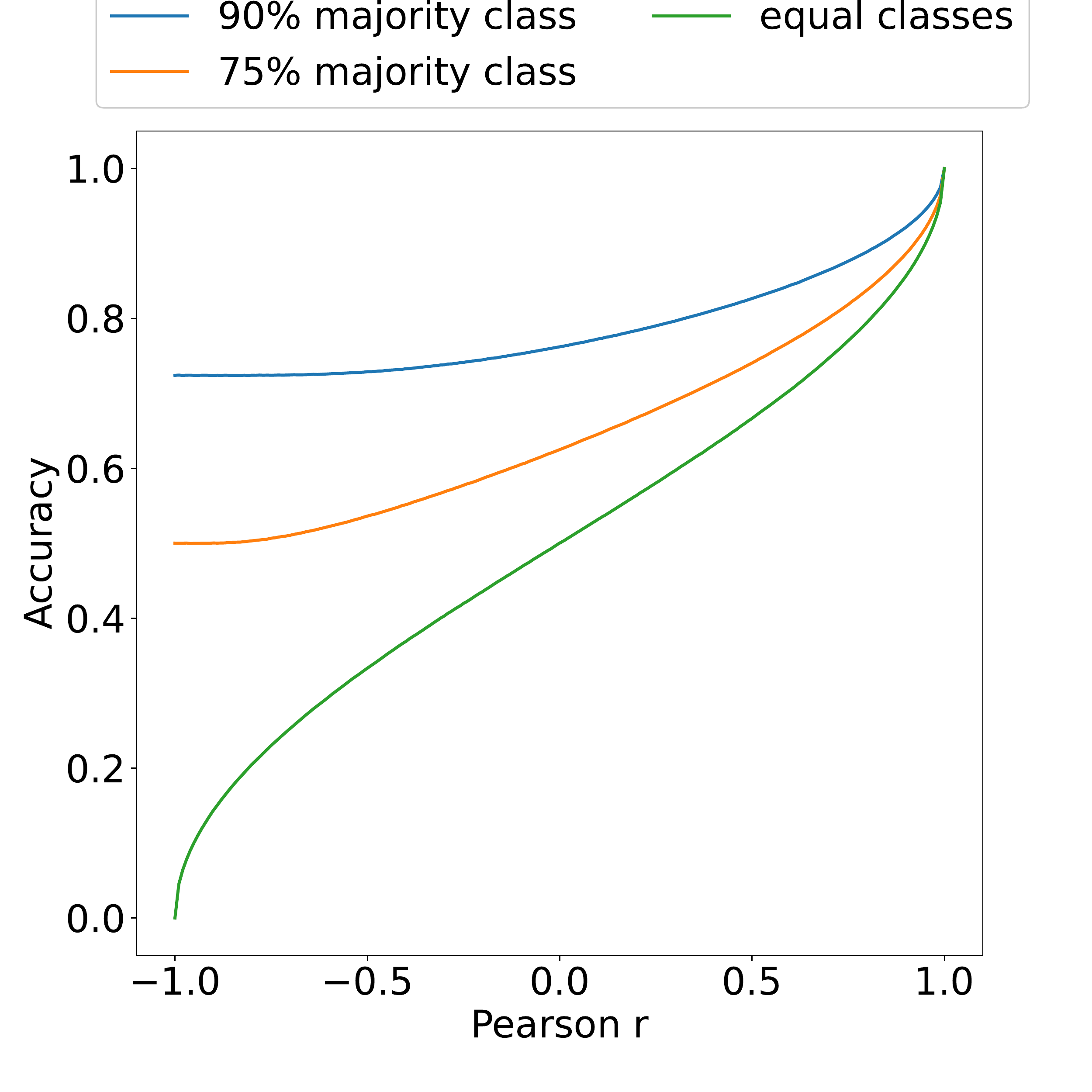}
\caption{}
\end{subfigure}
\hfill
\begin{subfigure}[t]{.475\textwidth}
\includegraphics[width=\linewidth]{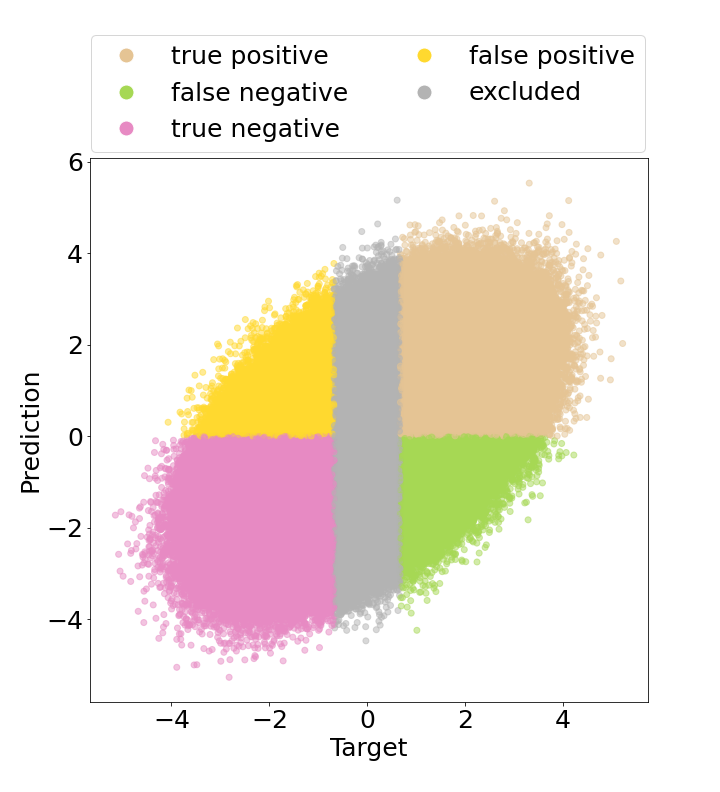}
\caption{}
\end{subfigure}
\hfill
\caption{a) correlation to accuracy transform for several majority class sizes. b) An example of simulated data discretization for evaluating extreme trait classification. Correlation parameter is 0.5. Examples between target quartiles are excluded.}
\label{figure:performance_conversion}
\end{figure}

As a result, we built a matrix with rows corresponding to correlation parameters, and columns corresponding to quality metrics. To convert metric A into metric B, we searched for a value of A closest to the given one and took a value of B from the same row. The conversion procedure is illustrated with some examples in \autoref{figure:performance_conversion}.

When converting estimates that come from binary classification of trait-based attributes, we decided to assume zero splits (although some class imbalance is likely). For MBTI, classes substantially differ in size. Thus, we use proportions as reported in the papers and exclude studies that do not provide them.

\subsection{Compilation of performance estimates}
\label{subsection:estimates}

In 2013, analysis of the results of the Workshop on Computational Personality Recognition (Shared Task) \cite{celli2013workshop} led the organizers to a  conclusion that a promising way to boost personality prediction performance is to perform feature selection over a very large feature space. That said, training algorithms with many tunable parameters carries a substantial risk of overfitting and thus evaluation requires extreme caution. Usually, overfitting considerations are crucial in machine learning (\cite{hastie2009elements}). With sufficiently complex models, it is possible to perfectly fit random labels, yet the result would be of no use for new examples (\cite{zhang2016understanding}).  Even the estimates on held-out data can be easily invalidated by reusing test examples (\cite{Gossmann2018TestDR,rao2008dangers}) or by including preprocessing procedures that rely on the entire dataset (\cite{hastie2009elements}). 
Despite this,  \cite{marengo2020digital} found that six out of 23 personality prediction studies included in their meta-analysis did not rely on held-out data for performance evaluation, and at the same time they found no effect of using a hold-out dataset on prediction performance.

Considering this, we decided to use a stricter criterion to select studies for performance comparison. We identified 44 papers (about 20\% of the sample) that explicitly state separating train, validation, and test phases (using separate sets, nested cross-validation, or any intermediate approach). Sixteen of these 44 papers only report metrics based on MAE or RMSE (13 are based on the first impression dataset --- \cite{Nurek2020HawkesmodeledTP,Li2020CRNetAD, Hernndez2020AudiovisualDL, Aslan2019MultimodalVP, Aslan2019MultimodalVA, Principi2019OnTE, Anselmi2019GenuinePR, Yu2019SpeakingSB, Xue2018DeepLP, Kampman2018InvestigatingAV, Escalante2018ExplainingFI, Kaya2018MultimodalPT, Wei2018DeepBR, Gltrk2018MultimodalFI, Bekhouche2017PersonalityTA, Wicaksana2017LayeredRA}), and thus we could not perform the conversion. We further excluded one study, in which the data preprocessing procedure depended on the entire dataset (\cite{Li2019QuantitativePP}). One paper we excluded as a duplicate (there was another paper by the same authors on the same dataset with identical quality estimates). We finally excluded from the set two MBTI prediction studies for which the information about class distributions is not available (\cite{Cui2017SurveyAO, Buraya2018MultiviewPP}). The final set consists of 24 studies predicting personal traits in varying combinations, including the Big-5 traits (\autoref{table:performance}), intelligence (\autoref{table:performance_intelligence}), values (\autoref{table:performance_values}), morals (\autoref{table:performance_other}) and the MBTI types (\autoref{table:performance_mbti}). 

Note that the use of separate validation set by itself is neither necessary nor sufficient to eliminate overfitting. It is possible to avoid the problem completely by setting all the parameters by hand, as in \cite{Kleanthous2016DetectingPT}. It is also possible to overfit to both validation and test set. However, it is easy to check whether or not the validation set use is reported and it is a good indicator of the fact that considerable efforts had been put to avoid overfitting. 

We decided to follow \cite{Azucar2018PredictingTB, marengo2020digital} in using Pearson correlation as a reference metric and added binary classification accuracy whenever we assumed equal class sizes. 

Several of the Pearson correlation estimates that emerged from the metric conversion procedure are negative. A possible reason for this is a significant violation of class equality assumption on which we relied.

In most cases, the Pearson correlation estimates for self-reported personality prediction are limited by 0.4--0.5. Quality estimates from three studies stand out. In \cite{Hall2017AmIW}, correlations exceed 0.6 for all the Big-5 traits. Estimates from  \cite{Subramanian2018ASCERTAINEA} suggest correlation about 0.75 between predicted and self-reported neuroticism. Finally, the estimated correlation for extraversion prediction in \cite{Salminen2020EnrichingSM} reaches 0.54.

First two studies are, however, in the grey zone regarding our selection criteria. In both of them, the classification is preceded by feature space transform and it is not clear whether it is limited to the train set or is based on the entire dataset. If the latter is the case, estimates could be exaggerated. There is large variation among the quality estimates for the last study, which could indicate a possible distortion from unaccounted effects.

There is also one estimate above 0.5 for vocabulary prediction  achieved on a private dataset in \cite{Titov2019FullscalePP}, yet there are too few estimates for the same trait to establish a reliable benchmark. 

There are fewer estimates for apparent personality prediction, but they tend to be higher than those of self-reported personality. This effect is very pronounced in studies that address both tasks on the same dataset (\cite{Segalin2017ThePW,Segalin2017SocialPT}). The same effect can be seen in an older study, \cite{Mairesse2007UsingLC}. The higher quality of apparent personality prediction may be due to the fact that annotators and prediction algorithms usually share the same source of the information. 

 Notably, perfect classification quality is reported for three of the Big-5 traits in \cite{Suhartono2017PersonalityPB}, despite the fact that even traits assessed by repeating the same test do not show perfect agreement.

\section{Conclusions and discussion}
\label{section:conclusions}

Personality inference based on digital data aspires to be a full-fledged alternative to conventional psychological tests. This implies that the automatic assessment should be able to identify a set of distinctive personal characteristics. Even more importantly, these characteristics should be consistently predictive of certain aspects of future behaviour. In this study, we have attempted to outline the progress made towards this task by looking at the solutions from the literature and reviewing various aspects of their performance from the perspective of the known psychometric instrument requirements.

The bulk of the available estimates refer to the degree of agreement of the predicted characteristics to the gold standard labels. These labels come from using conventional self-assessment tests or external annotators. Determining a single benchmark is complicated by the variety of agreement metrics used, and these metrics sometimes lack details necessary for their exact definition. Moreover, we have been able to identify explicit separation of train, validation, and testing phases in only approximately one in five studies, so there is a high risk of encountering estimates exaggerated due to overfitting. 
To approach compiling a benchmark, we have performed a screening based on validation procedures and approximately converted the reported quality estimates to Pearson correlations.  In most studies, the estimated correlation between self-reported and predicted personality traits is below 0.5. Several works report results that are higher by a substantial margin, but there is some methodological ambiguity, so some additional analysis is needed to find out whether the performance increase is due to some promising modelling decisions or measurement artifacts. The highest correlation estimates in the rest of the selected studies are achieved for extraversion and openness (0.48). 
Agreement estimates are consistently higher for apparent personality evaluation. They reach 0.7 or even 0.74 and 0.83 in some extraversion prediction cases. A probable reason is that reliance on the same digital data by machine learning algorithms and annotators results in shared sources of variation, while self-reported measures reflect a different kind of knowledge.

It is not straightforward to conclude what the correlations near 0.5 for predicted and self-reported traits mean qualitatively. Correlations between same-name traits measured with different Big-5 instruments are usually above 0.6 and occasionally reach 0.9 (\cite{john2008paradigm, soto2017next, ipip_comparison}). At the same time, correlations between same-name traits measured with instruments that come from different conceptual models can sometimes fall below 0.4 (\cite{pace2010similar}).

There is a single study we have found that has specifically raised the question of a sufficient degree of agreement of separate measures (\cite{carlson2012understanding}). The authors have investigated how correlation with an external outcome variable (for example, work satisfaction) may differ between a trait variable and its proxy (for example, extraversion measured with two distinct instruments) depending on the agreement between the two measures.  The authors pointed out that, in theory, a substantial difference in correlation with an outcome variable between trait and its proxy is possible even when trait and its proxy are closely correlated (the absolute difference can reach 0.29 for a trait-proxy correlation of 0.95).
The authors have also collected correlation estimates from the literature to get a real-world view on the same effect. As a conclusion, they recommend avoiding trait-proxy correlations below 0.5. However, the effects visible in the diagrams provided are not as grim as the theory suggests. For example, in 45\% of the cases with a trait-proxy correlation of 0.3, the difference in correlations with the outcome is below 0.1.

\begin{figure}

\centering

\begin{subfigure}[t]{.475\textwidth}
\centering
\includegraphics[width=\linewidth]{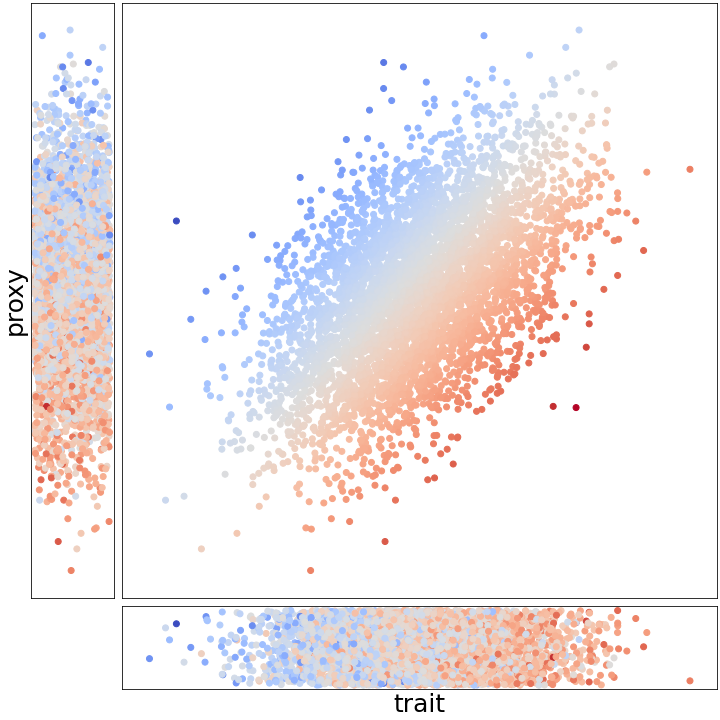}
\caption{trait-outcome correlation = 0.5, proxy-outcome correlation = -0.5}
\end{subfigure}
\hfill
\begin{subfigure}[t]{.475\textwidth}
\centering
\includegraphics[width=\linewidth]{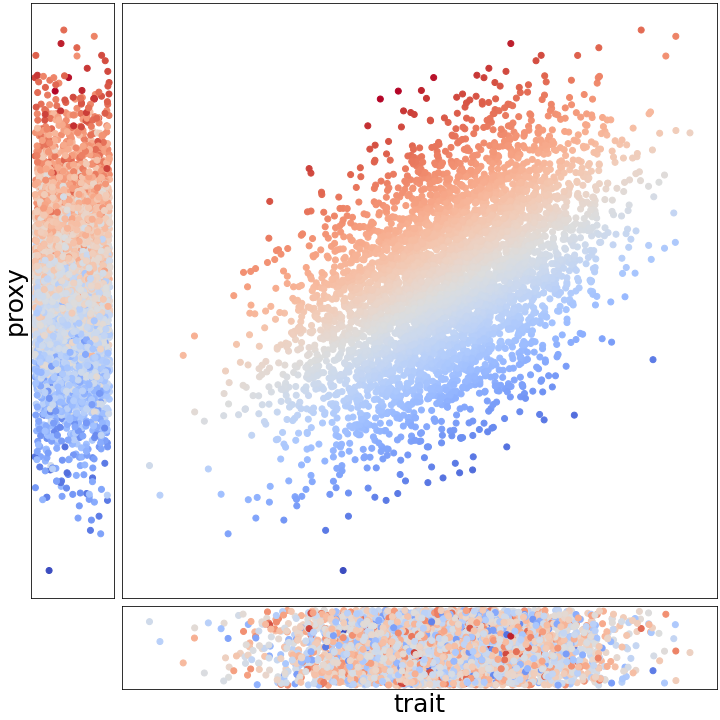}
\caption{trait-outcome correlation = 0.0, proxy-outcome correlation $\approx$ 0.866}
\end{subfigure}

 \medskip
\begin{subfigure}[t]{.475\textwidth}
\centering
\includegraphics[width=\linewidth]{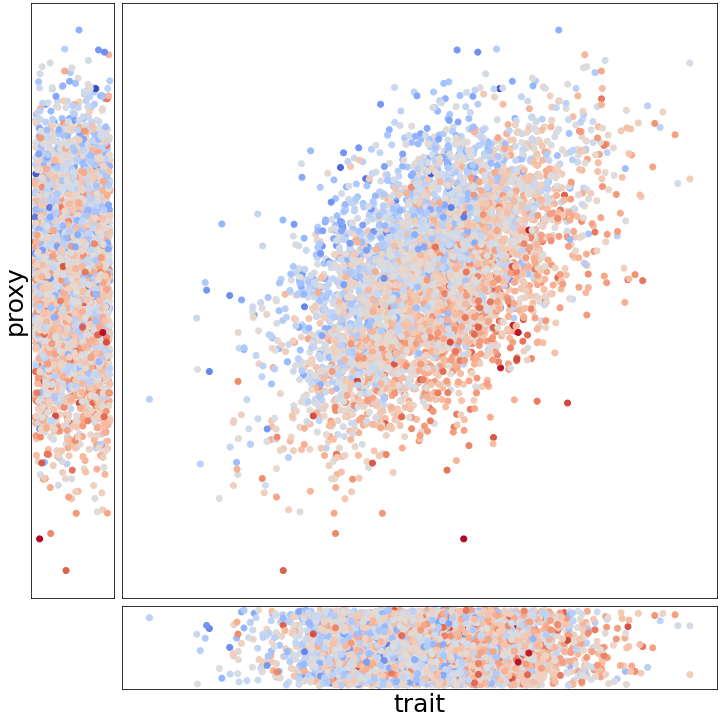}
\caption{trait-outcome correlation = 0.25, proxy-outcome correlation = -0.25}
\end{subfigure}
\hfill
\begin{subfigure}[t]{.475\textwidth}
\centering
\includegraphics[width=\linewidth]{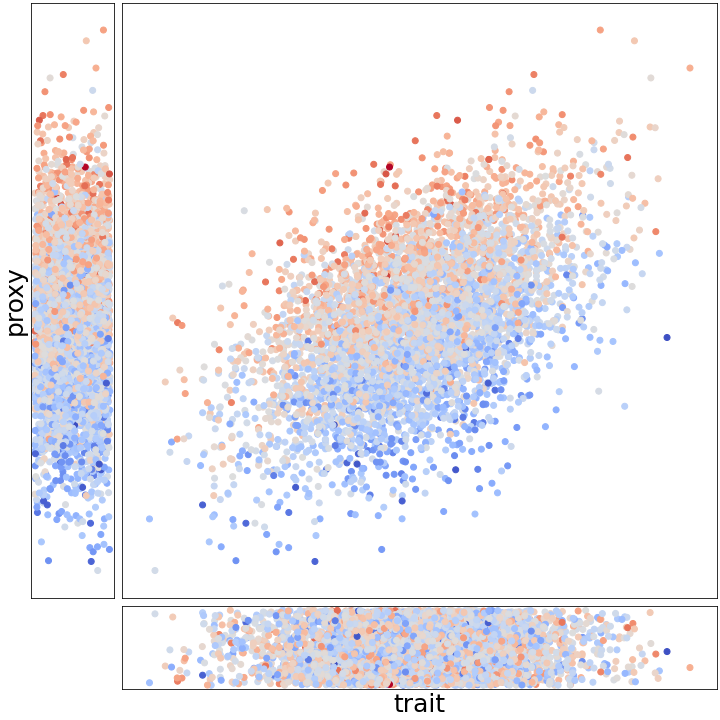}
\caption{trait-outcome correlation = 0.0, proxy-outcome correlation = 0.5}
\end{subfigure}
\hfill
\caption{Simulated examples of large discrepancy between correlations of two measures of the same characteristic (trait and proxy) with a separate outcome variable. Each point corresponds to a single respondent: spatial dimensions are trait and proxy values, and colour dimension displays outcome variable. Side scatter plots illustrate how outcome variable is related to a single dimension (additional dummy dimension is used to reduce point overlap). Correlation between trait and proxy is equal 0.5 for all examples. (a,b) correspond to worst case differences. Examples (c,d) correspond to lower, but still substantial magnitude of discrepancy. }
\label{figure:worst_case_corr}
\end{figure}

Still, the issue of large possible discrepancies in correlation with outcome variables needs some attention. We have explored it using some simulated examples for the case of two measures of a characteristic (trait and proxy) having a correlation of 0.5. Several possible joint distributions of trait, proxy, and outcome variable with substantial discrepancies in correlations between characteristic measures and the outcome variable are depicted in \autoref{figure:worst_case_corr}. The demonstrated distributions have a very particular structure: in each example, we can see that the outcome variable depends on a linear difference of trait and proxy variables. For this to occur in reality, the difference between the trait and its proxy must correspond to some characteristic that varies within the population, rather than random measurement error. The same characteristic must also be closely related to the chosen outcome variable. Going back to the problem of automatic personality prediction, we should expect large differences in the correlations with the outcome between the prediction and the original trait if the prediction error is largely systematic and the particular outcome variable is related to the error term.

The alternative psychometric validation procedures could provide a valuable  support in finding sources of possible systematic error, although, at the moment, the evidence is very limited. 

The content validity --- the requirement for the indicators that make up the prediction to be theoretically justifiable --- is hard to apply to machine learning-based solutions due to their data-driven nature. However, some researchers pay attention to the interpretability of the models, and there is already a substantial amount of knowledge about indicators that contribute to trait prediction. To use them effectively to assess prediction quality, we need to be able to clearly distinguish intuitive and counterintuitive dependencies. One study in our sample of papers (\cite{Lynn2020HierarchicalMF}) included and evaluation of an agreement between expert psychologists and models on feature importance. The agreement was only achieved for some of the traits. In all other studies, visualization and presentation of triggered features were at most accompanied by their post hoc discussion. One possible way to alleviate the post hoc interpretation problem is to build a lexicon of features that are predictive of personality traits across studies --- researchers from a related field have recently performed similar work by aggregating evidence on correlations of objective behavioural features and depressive mood symptoms (\cite{rohani2018correlations}). However, at the moment, there is little we can say about the content validity of inferred personality. 

It is rare to report factor structure and relationships of predicted characteristics in automatic personality prediction studies, although it is a critical part of theoretical propositions on personality. Available evidence suggests that the effective dimensionality of the predicted traits is lower than those of self-reported personality. At the same time, predicted personality traits tend to retain the general correlation pattern. 

Furthermore, there are very modest results on the cross-domain generalizability of personality prediction models in the sample of papers included in the analysis. Examinations of prediction stability (reliability) are too rare to make any conclusions. There is surprisingly little research on the stability of the estimates, considering that it does not require annotated data. 

Finally, criterion validity ---  the requirement for a trait to have theoretically justifiable relations with the external outcome variables --- is essential for the predicted personality profile to be useful. So far, few attempts were made to examine the relationship between the predicted traits and the external features. There is some evidence that inferred personality is predictive of an occupation,  preference for a particular brand, and some other psychological characteristics and decisions of a person. However, theoretical expectations are usually not clarified, and it is possible to argue for an alternative explanation of the predictability through common bias rather than an accurate personality profile. 

Thus, personality prediction using machine learning is at a relatively early stage as a field of research. It is possible to compile a benchmark for prediction quality, but existing evidence is insufficient to set clear performance expectations for real-world tasks. We do not know in advance which properties of the original characteristics will be exhibited by the predicted traits (for example, a relation  between predicted extraversion and sociability cannot be taken for granted). In fact, we cannot be sure that the predicted characteristics are consistent with the understanding of personality traits from psychological science.  Thus, we think that additional investigation of prediction properties, perhaps inspired by the psychometric validation procedures, is desperately needed. Meanwhile, researchers have built up a decent amount of personality-annotated data available for research. That can facilitate future progress in the field by both relieving new projects from the labour-intensive obligatory data collection phase and making new results more comparable.

\section{Limitations}
The present study has several limitations. First, there are sample limitations. The literature aggregation process has been stopped when most expansion iterations started to give no new results. Consequently, there are still many studies in the sample that have not been used for expansion. Another possible way to extend the sample is to employ some of the other search engines, or take into account some of the studies published prior to 2017. 

Second, parameter extraction from the studies has been performed by one researcher, so there are likely some mistakes in parameter identification. 

Third, most works from the sample are devoted to prediction of the Big-5 traits, and there are not enough studies to make reliable quality estimates for traits based on other personality models. This is probably  a result of an asymmetry of the general set of studies, but it could be possible to collect more studies by focusing specifically on alternative models of personality. 

Finally, there is some ambiguity regarding the role of several studies with unusually high prediction performance estimates in conclusions on achieved prediction quality. It can potentially be addressed by attempting to reproduce the results using available datasets. 

\section*{Acknowledgements}
We would like to offer special thanks to Yaroslav Golubev for reading, editing, and providing valuable advice on the article. We have received useful and constructive recommendations when discussing results with Yanina Ledovaya, Sergey Titov, and Dmitry Abbakumov. We are particularly grateful to Maxim Katinov and Sergey Yakubanets for creating and supporting our research team.

\begin{landscape}
\captionsetup{width=20cm}
\begin{longtable}{|p{6cm}|p{1.2cm}|p{8cm}|p{2.7cm}|p{.7cm}|p{1.1cm}|}
\caption{Datasets annotated with personality. The column \textit{annotation} contains personality instruments used for dataset annotation. For the Big-5, we include the exact instrument where possible.
The label \textit{other} indicates the use of additional psychometric instruments (neither Big-5 nor MBTI). The Big-5 instruments are: BFI-44 (\cite{john2008paradigm}), NEO-PI-R (\cite{costa1992revised}), NEO-FFI (\cite{costa1992revised}), BFI-10 (\cite{rammstedt2007measuring}), BFI-2 (\cite{soto2017next}), BFQ (\cite{caprara1993big}), BFMS (\cite{perugini2002analyzing}), IPIP (\cite{ipip_project, goldberg2006international}), TIPI (\cite{gosling2003very}). The column \textit{label type} distinguishes self-reported (S) personality, apparent personality annotated by unacquainted judges (A), and mined personality labels that users had shared in social media (M). The last column indicates whether each person corresponds to a single annotated item. }
\label{table:datasets}\\
\hline
dataset & size & data & psychometric instruments
 & label type & item is person \\
\hline
\endfirsthead
\hline
dataset & size & data & psychometric instruments
 & label type & item is person \\
\hline
\endhead

myPersonality (full) (\cite{my_personality, Kosinski2013PrivateTA}) & $\ge$115864 & Facebook likes, status updates, friendship graph, etc & IPIP, other &  S & \checkmark \\
\hline

TwiSty (\cite{Verhoeven2016TwiStyAM}) & 18168 & tweets & MBTI &  M & \checkmark \\
\hline
PANDORA (\cite{Gjurkovic2020PANDORATP}) & 10288 & Reddit comments & MBTI, Big-5 &  M & \checkmark \\
\hline
first impressions (\cite{ponce2016chalearn}) & 10000 & vlog videos & Big-5 &  A & $\times$ \\
\hline
first impressions v2 (\cite{escalante2018explaining}) & 10000 & vlog videos & Big-5 &  A & $\times$ \\
\hline
MBTI9k (\cite{gjurkovic2018reddit}) & 9111 & Reddit comments and activity features & MBTI &  M & \checkmark \\
\hline
Kaggle-MBTI (\cite{kaggle_mbti}) & 8675 & PersonalityCafe forum posts & MBTI &  M & \checkmark \\
\hline
SOC essays (\cite{pennebaker1999linguistic} or \cite{argamon2005lexical}) & 2479 & Stream-of-consciousness essays & BFI-44 or NEO-FFI &  S & ? \\
\hline
Sina Micro-blog (\cite{Liu2016DeepLF}) & 1552 & Sina micro-blog text & BFQ &  S & \checkmark \\
\hline
Twitter-MBTI-1500 (\cite{Plank2015PersonalityTO}) & 1500 & tweets & MBTI &  M & \checkmark \\
\hline
b5 (\cite{Ramos2018BuildingAC}) & 1082 & Facebook posts \& other text & BFI-44 &  S & \checkmark \\
\hline
Personal-ITY (\cite{Bassignana2020PersonalITYAN}) & 1048 & Youtube comments & MBTI &  M & \checkmark \\
\hline
BIT Speaker Personality Corpus (\cite{Zhang2017SocialPE}) & 1031 & speech clips & NEO-PI-R &  A & $\times$ \\
\hline
PAN-2015 (\cite{rangel2015overview}) & 726 & tweets & BFI-10 &  S & \checkmark \\
\hline
CLiPS (\cite{verhoeven2014clips}) & 697 & essays & BFI-2, MBTI &  S & \checkmark \\
\hline
SPC (\cite{Mohammadi2012AutomaticPP}) & 640 & speech clips & BFI-10 &  A & $\times$ \\
\hline
PhoneStudy (\cite{stachl2020predicting}) & 624 & mobile phone activity data & BFSI &  S & \checkmark \\
\hline
RusNeuroPsych (\cite{Litvinova2018RusNeuroPsychOC}) & 455 & stream-fo-consicousness essays & NEO-FFI, other &  S & \checkmark \\
\hline

youtube (\cite{biel2013youtube}) & 442 & vlog videos & TIPI &  A & \checkmark \\
\hline
TxPI-u (\cite{RamrezdelaRosa2018TxPIuAR}) & 417 & handwritten essays with transcription & TIPI &  S & \checkmark \\
\hline
youtube-transcribed (\cite{biel2013hi}) & 408 & text transcriptions of vlog videos & TIPI &  A & \checkmark \\
\hline
Psycho-Flickr (\cite{cristani2013unveiling}) & 300 & favorite pictures from Flickr & BFI-10 &  S\&A & \checkmark \\
\hline
LIWC-facebook (\cite{Hall2017AmIW}) & 282 & LIWC (\cite{tausczik2010psychological}) features extracted from facebook texts & BFI-44 &  S & \checkmark \\
\hline
CXD (\cite{levitan2015cross}) & 278 & audio-conversations & NEO-FFI &  S & \checkmark \\
\hline
myPersonality-small (\cite{celli2013workshop}) & 250 & Facebook status updates and social network measures & IPIP &  S & \checkmark \\
\hline
netsense (\cite{meng2014analyzing}) & 200 & mobile phone activity data & BFI-44 &  S & \checkmark \\
\hline
Personae (\cite{Luyckx2008PersonaeAC}) & 145 & essays & MBTI &  S & \checkmark \\
\hline
EAR (\cite{Mehl2006PersonalityII}) & 96 & sample of daily life sound recordings with transcriptions & BFI-44 &  S\&A & \checkmark \\
\hline
twitter-egyptian (\cite{Salem2019PersonalityTF}) & 92 & tweets \& Twitter profile data & NEO-PI-R &  S & \checkmark \\
\hline
ELEA (\cite{sanchez2013emergent}) & 85 & audio- and videorecordings of group meetings & NEO-FFI, other &  S & \checkmark \\
\hline
Ascertain (\cite{Subramanian2018ASCERTAINEA}) & 58 & multimodal sensor recordings & BFMS &  S & \checkmark \\
\hline
friends\&family (\cite{aharony2011social}) & 53 & mobile phone activity data & BFI-44, other &  S & \checkmark \\
\hline
MULTISIMO (\cite{koutsombogera2018modeling}) & 49 & audio, video \& depth recordings & BFI-44 &  S & \checkmark \\
\hline
StudentLife (\cite{wang2014studentlife}) & 48 & mobile phone activity data & BFI-44 &  S & \checkmark \\
\hline
personality-nonsocial (\cite{dotti2018behavior}) & 46 & 3d joint trajectories \& depth images & BFI-10 &  S & \checkmark \\
\hline
MAPTRAITS 2014 (\cite{celiktutan2014maptraits}) & 44 & videoclips & Big-5 &  A & $\times$ \\
\hline
Amigos (\cite{correa2018amigos}) & 40 & physiological sensors & BFMS, other &  S & \checkmark \\
\hline
Vyaktitv (\cite{Khan2020VyaktitvAM}) & 38 & video/audio recordings of dialogues, transcriptions & IPIP &  S & \checkmark \\
\hline
Wearable sensors (\cite{Butt2020MultimodalPT}) & 28 & physiological sensors & Big-5 &  S & \checkmark \\
\hline
SALSA (\cite{alameda2015salsa}) & 18 & recordings from cameras and wearable sensors & BFI-44 &  S & \checkmark \\
\hline
mhhri (\cite{celiktutan2017multimodal}) & 18 & multimodal recordings of human-robot interactions & BFI-10, other &  S & \checkmark \\
\hline
\end{longtable}

\captionsetup{width=20cm}
\begin{longtable}{|p{2.1cm}|p{1.4cm}|p{1.4cm}|p{1.4cm}|p{1.4cm}|p{1.4cm}|p{4.4cm}|p{.9cm}|p{.8cm}|}
 \caption{Big-5 prediction performance (Pearson correlation/binary accuracy). Whenever a study reports several results on the same data, we include the highest result. If a study includes results on several subsets of the data, as well as a superset of them, we only include the latter. Asterisk marks the estimates that either come directly from the papers or are calculated using exact formulas.
If the results are achieved on one of the reusable corpora, we mention its name. Otherwise, we mark the dataset as private and include the data type (SN stands for social network). Datasets are SOC essays (\cite {pennebaker1999linguistic, argamon2005lexical}), myPersonality (small) (\cite{celli2013workshop}), youtube (text) (\cite{celli2014workshop}), myPersonality (full) (\cite{my_personality, Kosinski2013PrivateTA}), PhoneStudy (\cite{stachl2020predicting}), TxPI-u (\cite{RamrezdelaRosa2018TxPIuAR}), ASCERTAIN (\cite{Subramanian2018ASCERTAINEA}), PsychoFlickr (\cite{cristani2013unveiling}), ELEA (\cite{sanchez2013emergent}), BIT corpus (\cite{Zhang2017SocialPE}), LIWC-facebook (\cite{Hall2017AmIW}).
The last column distinguishes self-reported (checkmark) and apparent (cross) labels. Cell shading indicates the magnitude of the correlation.}
\label{table:performance}\\
    \hline 
    Study &  O & C & E & A & N & data & year & self-report \\
     \hline
     \endfirsthead
         \hline 
    Study &  O & C & E & A & N & data & year & self-report \\
     \hline
     \endhead
\cite{Lynn2020HierarchicalMF} & \cellcolor[rgb]{0.9604900613294117,0.6162764239411764,0.49546660049411767}$.48^*$/.66 & \cellcolor[rgb]{0.9692885689999999,0.6849817470823529,0.5689753262588235}$.39^*$/.63 & \cellcolor[rgb]{0.966922105,0.6519686451529412,0.5319971156235295}$.43^*$/.64 & \cellcolor[rgb]{0.968105337,0.6684751961176472,0.5504862209411766}$.41^*$/.63 & \cellcolor[rgb]{0.968105337,0.6684751961176472,0.5504862209411766}$.41^*$/.63 & myPersonality (full) & 2020 & \checkmark\\
\hline
\cite{RoblesGranda2020JointlyPJ} & \cellcolor[rgb]{0.9527607176705882,0.7829647976,0.6986457713058823}.22/.57 & \cellcolor[rgb]{0.9545658223607844,0.7790545618666667,0.6925308598078431}.23/.57 & \cellcolor[rgb]{0.9648353582352941,0.7446136745882352,0.6432388753529412}.29/.59 & \cellcolor[rgb]{0.9193759889058823,0.8312727235294118,0.7828736304470588}.11/.54 & \cellcolor[rgb]{0.8674276350862745,0.864376599772549,0.8626024620196079}.00/.50 & private (multimodal) & 2020 & \checkmark\\
\hline
\cite{Salminen2020EnrichingSM} & \cellcolor[rgb]{0.963806056435294,0.6341884145294118,0.5137208491529413}.46/.65 & \cellcolor[rgb]{0.925563423,0.8255172980705883,0.7711363078117647}.13/.54 & \cellcolor[rgb]{0.9593847296274509,0.6103057604117648,0.4893818509411765}.49/.66 & \cellcolor[rgb]{0.9675442976352941,0.7308497161882352,0.6246854782352941}.32/.60 & \cellcolor[rgb]{0.839351442772549,0.861166825654902,0.8944937634156863}-.05/.48 & SOC essays & 2020 & \checkmark\\
\hgrayline
\arrayrulecolor{lightgray}\hline\arrayrulecolor{black}
 & \cellcolor[rgb]{0.9563709270509804,0.7751443261333334,0.6864159483098039}.24/.58 & \cellcolor[rgb]{0.6933212848235294,0.7963141317058823,0.9863077805294118}-.29/.41 & \cellcolor[rgb]{0.9476541841529411,0.5659764341686274,0.4474781480392157}.54/.68 & \cellcolor[rgb]{0.9434315296666667,0.8022762536156862,0.7291715979137255}.18/.56 & \cellcolor[rgb]{0.5977767754941177,0.7273297248823529,0.9997767317764705}-.43/.36 & myPersonality (small) &  & \checkmark\\
\hgrayline
 & \cellcolor[rgb]{0.9458543787882353,0.5595649564588235,0.44151331917647063}.55/.69 & \cellcolor[rgb]{0.839351442772549,0.861166825654902,0.8944937634156863}-.05/.48 & \cellcolor[rgb]{0.921406221227451,0.49142041718431373,0.38340843537647057}.63/.72 & \cellcolor[rgb]{0.7727059486039215,0.8389782172392156,0.9493187599137255}-.17/.45 & \cellcolor[rgb]{0.6881884831921569,0.7931783792980391,0.9880381043568628}-.30/.40 & youtube (text) &  &  $\times$\\
    \hline
     \cite{Stachl2020PredictingPF} & \cellcolor[rgb]{0.9648353582352941,0.7446136745882352,0.6432388753529412}$.29^*$/.59 & \cellcolor[rgb]{0.9669624381411764,0.7356700026196078,0.6308765394784314}$.31^*$/.60 & \cellcolor[rgb]{0.9698511524117647,0.6958300595294117,0.5813117740784314}$.37^*$/.62 & \cellcolor[rgb]{0.8918168921215687,0.8519732770431372,0.829085274254902}$.05^*$/.52 & \cellcolor[rgb]{0.9193759889058823,0.8312727235294118,0.7828736304470588}$.11^*$/.54 & PhoneStudy & 2020 & \checkmark\\
     \hline
     \cite{Rahman2019PersonalityDF} & \cellcolor[rgb]{0.9658988981882353,0.7401418386039216,0.6370577074156862}.30/.60 & \cellcolor[rgb]{0.8995432066000001,0.8475002359999999,0.8177890744}.07/.52 & \cellcolor[rgb]{0.6672529243333334,0.7791764569999999,0.992959213}-.33/.39 & \cellcolor[rgb]{0.8006008472941177,0.8503583215607843,0.9300075603921568}-.12/.46 & \cellcolor[rgb]{0.9595176584705882,0.7669728545098039,0.6741447150392157}.25/.58 & SOC essays & 2019 & \checkmark\\
\hline
\cite{Titov2019FullscalePP} & \cellcolor[rgb]{0.9473454036,0.7946955048,0.7169905058}.20/.56 & \cellcolor[rgb]{0.945540298909804,0.7986057405333333,0.7231054172980392}.19/.56 & \cellcolor[rgb]{0.9648353582352941,0.7446136745882352,0.6432388753529412}.29/.59 & \cellcolor[rgb]{0.925563423,0.8255172980705883,0.7711363078117647}.13/.54 & \cellcolor[rgb]{0.9357737696666666,0.8122367012392158,0.7471564735843139}.16/.55 & SN profile & 2019 & \checkmark\\
\hgrayline
 & \cellcolor[rgb]{0.9675442976352941,0.7308497161882352,0.6246854782352941}.32/.60 & \cellcolor[rgb]{0.968105337,0.6684751961176472,0.5504862209411766}.41/.63 & \cellcolor[rgb]{0.9685329496823529,0.7158412919058823,0.6060967478823529}.34/.61 & \cellcolor[rgb]{0.8877520159490196,0.8540404974980391,0.8346714722156863}.04/.51 & \cellcolor[rgb]{0.9563709270509804,0.7751443261333334,0.6864159483098039}.24/.58 & SN subscriptions &  & \checkmark\\
 \hline
    \cite{nave2018musical} & \cellcolor[rgb]{0.9658988981882353,0.7401418386039216,0.6370577074156862}$.30^*$/.60 & \cellcolor[rgb]{0.9434315296666667,0.8022762536156862,0.7291715979137255}$.18^*$/.56 & \cellcolor[rgb]{0.9491505082901961,0.7907852690666667,0.7108755943019608}$.21^*$/.57 & \cellcolor[rgb]{0.9383263563333333,0.8089165520313726,0.741161515027451}$.17^*$/.55 & \cellcolor[rgb]{0.9357737696666666,0.8122367012392158,0.7471564735843139}$.16^*$/.55 & myPersonality (full) & 2018 & \checkmark\\
     \hline
          
          \cite{RamrezdelaRosa2018OverviewOT} & \cellcolor[rgb]{0.8674276350862745,0.864376599772549,0.8626024620196079}.00/.50 & \cellcolor[rgb]{0.8995432066000001,0.8475002359999999,0.8177890744}.07/.52 & \cellcolor[rgb]{0.9094595977529412,0.8393864797647058,0.8003313524235294}.09/.53 & \cellcolor[rgb]{0.8513716370313725,0.8631253107372548,0.8810638238392158}-.03/.49 & \cellcolor[rgb]{0.8796222636039216,0.8581749384078431,0.845843868137255}.03/.51 & TxPI-u (image) & 2018 & \checkmark\\
\hgrayline
 & \cellcolor[rgb]{0.8918168921215687,0.8519732770431372,0.829085274254902}.05/.52 & \cellcolor[rgb]{0.8433581741921568,0.8618196540156863,0.8900171168901961}-.04/.49 & \cellcolor[rgb]{0.8918168921215687,0.8519732770431372,0.829085274254902}.05/.52 & \cellcolor[rgb]{0.9127650614705882,0.8366818943529412,0.7945121117647058}.10/.53 & \cellcolor[rgb]{0.8918168921215687,0.8519732770431372,0.829085274254902}.05/.52 & TxPI-u (text) &  & \checkmark\\
 \hline
\cite{Subramanian2018ASCERTAINEA} & \cellcolor[rgb]{0.945540298909804,0.7986057405333333,0.7231054172980392}.19/.56 & \cellcolor[rgb]{0.9548534056117647,0.5916223450078432,0.4713374634901961}.51/.67 & \cellcolor[rgb]{0.9649113881372549,0.6401590780588234,0.5198055987058824}.45/.65 & \cellcolor[rgb]{0.9595176584705882,0.7669728545098039,0.6741447150392157}.25/.58 & \cellcolor[rgb]{0.8653913329372549,0.3711276720470588,0.2957689564156863}.75/.77 & ASCERTAIN & 2018 & \checkmark\\
\hline
 
         \cite{Sun2018WhoAI} & \cellcolor[rgb]{0.9595176584705882,0.7669728545098039,0.6741447150392157}.25/.58 & \cellcolor[rgb]{0.9685329496823529,0.7158412919058823,0.6060967478823529}.34/.61 & \cellcolor[rgb]{0.951253794882353,0.5787993895882353,0.4594078057647059}.53/.68 & \cellcolor[rgb]{0.9685329496823529,0.7158412919058823,0.6060967478823529}.34/.61 & \cellcolor[rgb]{0.9688625003647059,0.7108384838117647,0.5999005044313725}.35/.61 & youtube (text) & 2018 &  $\times$\\
\hgrayline
 & \cellcolor[rgb]{0.9473454036,0.7946955048,0.7169905058}.20/.56 & \cellcolor[rgb]{0.9281160096666666,0.8221971488627451,0.765141349254902}.14/.54 & \cellcolor[rgb]{0.9434315296666667,0.8022762536156862,0.7291715979137255}.18/.56 & \cellcolor[rgb]{0.9616447383764706,0.7580291825411765,0.6617823791647058}.27/.59 & \cellcolor[rgb]{0.9491505082901961,0.7907852690666667,0.7108755943019608}.21/.57 & SOC essays &  & \checkmark\\
 \hline

\cite{Arnoux201725TT} & \cellcolor[rgb]{0.9698511524117647,0.6958300595294117,0.5813117740784314}$.37^*$/.62 & \cellcolor[rgb]{0.968203399,0.7208441,0.6122929913333334}$.33^*$/.61 & \cellcolor[rgb]{0.9616447383764706,0.7580291825411765,0.6617823791647058}$.27^*$/.59 & \cellcolor[rgb]{0.9648353582352941,0.7446136745882352,0.6432388753529412}$.29^*$/.59 & \cellcolor[rgb]{0.9677109263333333,0.6629730124627451,0.5443231858352942}$.42^*$/.64 & private (tweets) & 2017 & \checkmark\\
\hline

\cite{Wei2017BeyondTW} & \cellcolor[rgb]{0.963806056435294,0.6341884145294118,0.5137208491529413}.46/.65 & \cellcolor[rgb]{0.963806056435294,0.6341884145294118,0.5137208491529413}.46/.65 & \cellcolor[rgb]{0.9604900613294117,0.6162764239411764,0.49546660049411767}.48/.66 & \cellcolor[rgb]{0.968105337,0.6684751961176472,0.5504862209411766}.41/.63 & \cellcolor[rgb]{0.968105337,0.6684751961176472,0.5504862209411766}.41/.63 & private (SN multimodal) & 2017 & \checkmark\\
\hline
         \cite{Segalin2017ThePW} & \cellcolor[rgb]{0.9476541841529411,0.5659764341686274,0.4474781480392157}.54/.68 & \cellcolor[rgb]{0.9318312966,0.5190855232,0.4064796086}.60/.70 & \cellcolor[rgb]{0.8921375427882353,0.4253887370980392,0.33328927276078435}.70/.75 & \cellcolor[rgb]{0.9343054549058823,0.525917511654902,0.4122864740431373}.59/.70 & \cellcolor[rgb]{0.908908026654902,0.46243263716862765,0.36095039415294133}.66/.73 & PsychoFlickr (apparent) & 2017 &  $\times$\\
\hgrayline
 & \cellcolor[rgb]{0.9595176584705882,0.7669728545098039,0.6741447150392157}.25/.58 & \cellcolor[rgb]{0.9527607176705882,0.7829647976,0.6986457713058823}.22/.57 & \cellcolor[rgb]{0.9627082783294117,0.7535573465568628,0.655601211227451}.28/.59 & \cellcolor[rgb]{0.9616447383764706,0.7580291825411765,0.6617823791647058}.27/.59 & \cellcolor[rgb]{0.9527607176705882,0.7829647976,0.6986457713058823}.22/.57 & PsychoFlickr (self-report) &  & \checkmark\\
 \hline

\cite{Kindiroglu2017MultidomainAM} & \cellcolor[rgb]{0.9226814526235294,0.8285681381176471,0.7770543897882353}.12/$.54^*$ & \cellcolor[rgb]{0.968203399,0.7208441,0.6122929913333334}.33/$.61^*$ & \cellcolor[rgb]{0.8204010983882353,0.2867649126352941,0.2451595198}\textcolor{white}{.83/$.81^*$} & \cellcolor[rgb]{0.9616447383764706,0.7580291825411765,0.6617823791647058}.27/$.59^*$ & \cellcolor[rgb]{0.968203399,0.7208441,0.6122929913333334}.33/$.61^*$ & ELEA & 2017 &  $\times$\\
         \hline
        \cite{ Asadzadeh2017AnalyzingFA} & \cellcolor[rgb]{0.9696829796666666,0.6904839307372549,0.5751383613647059}$.38^*$/.62 & \cellcolor[rgb]{0.9648353582352941,0.7446136745882352,0.6432388753529412}$.29^*$/.59 & \cellcolor[rgb]{0.9685329496823529,0.7158412919058823,0.6060967478823529}$.34^*$/.61 & \cellcolor[rgb]{0.9527607176705882,0.7829647976,0.6986457713058823}$.22^*$/.57 & \cellcolor[rgb]{0.9616447383764706,0.7580291825411765,0.6617823791647058}$.27^*$/.59 & myPersonality (full) & 2017 & \checkmark\\
        \hline
\cite{Guntuku2017StudyingPT} & \cellcolor[rgb]{0.9383263563333333,0.8089165520313726,0.741161515027451}$.17^*$/.55 & \cellcolor[rgb]{0.9434315296666667,0.8022762536156862,0.7291715979137255}$.18^*$/.56 & \cellcolor[rgb]{0.9527607176705882,0.7829647976,0.6986457713058823}$.22^*$/.57 & \cellcolor[rgb]{0.9545658223607844,0.7790545618666667,0.6925308598078431}$.23^*$/.57 & \cellcolor[rgb]{0.9677109263333333,0.6629730124627451,0.5443231858352942}$.42^*$/.64 & private (SN posted images) & 2017 & \checkmark\\
        \hline
        \cite{Segalin2017SocialPT} & \cellcolor[rgb]{0.925563423,0.8255172980705883,0.7711363078117647}.13/$.54^*$ & \cellcolor[rgb]{0.9357737696666666,0.8122367012392158,0.7471564735843139}.16/$.55^*$ & \cellcolor[rgb]{0.925563423,0.8255172980705883,0.7711363078117647}.13/$.54^*$ & \cellcolor[rgb]{0.925563423,0.8255172980705883,0.7711363078117647}.13/$.54^*$ & \cellcolor[rgb]{0.925563423,0.8255172980705883,0.7711363078117647}.13/$.54^*$ & PsychoFlickr (self-report) & 2017 & \checkmark\\
\hgrayline
 & \cellcolor[rgb]{0.9685329496823529,0.7158412919058823,0.6060967478823529}.34/$.61^*$ & \cellcolor[rgb]{0.9548534056117647,0.5916223450078432,0.4713374634901961}.51/$.67^*$ & \cellcolor[rgb]{0.9649113881372549,0.6401590780588234,0.5198055987058824}.45/$.65^*$ & \cellcolor[rgb]{0.966922105,0.6519686451529412,0.5319971156235295}.43/$.64^*$ & \cellcolor[rgb]{0.9440545734235294,0.5531534787490197,0.4355484903137255}.56/$.69^*$ & PsychoFlickr (apparent) &  &  $\times$\\
\hline

\cite{Siddique2017BilingualWE} & \cellcolor[rgb]{0.9440545734235294,0.5531534787490197,0.4355484903137255}.56/.69 & \cellcolor[rgb]{0.9649113881372549,0.6401590780588234,0.5198055987058824}.45/.65 & \cellcolor[rgb]{0.9566532109764706,0.598033822717647,0.4773022923529412}.50/.67 & \cellcolor[rgb]{0.963806056435294,0.6341884145294118,0.5137208491529413}.46/.65 & \cellcolor[rgb]{0.9593847296274509,0.6103057604117648,0.4893818509411765}.49/.66 & BIT corpus & 2017 &  $\times$\\
\hline

\cite{ Zhang2017SocialPE} & \cellcolor[rgb]{0.968105337,0.6684751961176472,0.5504862209411766}.41/$.63^*$ & \cellcolor[rgb]{0.9057834780117647,0.4551856921647059,0.35533588384705883}.67/$.73^*$ & \cellcolor[rgb]{0.8734022825529412,0.3869596390588235,0.3063324639764706}.74/$.77^*$ & \cellcolor[rgb]{0.9268829799882353,0.5054215462901961,0.3948658777137255}.61/$.71^*$ & \cellcolor[rgb]{0.8995343807254902,0.4406918021568627,0.34410686323529416}.68/$.74^*$ & BIT corpus & 2017 &  $\times$\\
        \hline
        \cite{Hall2017AmIW} & \cellcolor[rgb]{0.9244088216823529,0.49858955783529413,0.38905901227058826}$.62^*$/.71 & \cellcolor[rgb]{0.8995343807254902,0.4406918021568627,0.34410686323529416}$.68^*$/.74 & \cellcolor[rgb]{0.8921375427882353,0.4253887370980392,0.33328927276078435}$.70^*$/.75 & \cellcolor[rgb]{0.8995343807254902,0.4406918021568627,0.34410686323529416}$.68^*$/.74 & \cellcolor[rgb]{0.9268829799882353,0.5054215462901961,0.3948658777137255}$.61^*$/.71 & LIWC-facebook & 2017 & \checkmark\\
        \hline
\cite{Suhartono2017PersonalityPB} & \cellcolor[rgb]{0.7115538514588235,0.03333685421176469,0.1544847098509804}.99/$.95^*$ & \cellcolor[rgb]{0.7115538514588235,0.03333685421176469,0.1544847098509804}.99/$.95^*$ & \cellcolor[rgb]{0.705673158,0.01555616,0.150232812}1.00/$1.00^*$ & \cellcolor[rgb]{0.7115538514588235,0.03333685421176469,0.1544847098509804}.99/$.95^*$ & \cellcolor[rgb]{0.705673158,0.01555616,0.150232812}1.00/$1.00^*$ & private (SN profile \& text) & 2017 &  $\times$\\
\hline

\end{longtable}

\captionsetup{width=20cm}
\begin{longtable}{|p{2.5cm}|p{1.4cm}|p{1.4cm}|p{1.4cm}|p{1.4cm}|p{4cm}|p{1.5cm}|}
 \caption{MBTI prediction performance converted into Pearson correlation values. Cell shading indicates the magnitude of the correlation.}
\label{table:performance_mbti}\\

    \hline 
    Study &  E/I & N/F & T/S & J/P & data & year\\
    \hline
     \cite{Yamada2019IncorporatingTI} & \cellcolor[rgb]{0.963806056435294,0.6341884145294118,0.5137208491529413}0.46&
\cellcolor[rgb]{0.9660167198392157,0.6461297415882352,0.5258903482588235}0.44&
\cellcolor[rgb]{0.9660167198392157,0.6461297415882352,0.5258903482588235}0.44&
\cellcolor[rgb]{0.9595176584705882,0.7669728545098039,0.6741447150392157}0.25&
 private (SN activity) & 2019 \\
    \hline
\end{longtable}
\begin{longtable}{|p{2.5cm}|p{1.7cm}|p{1.7cm}|p{4cm}|p{1.5cm}|p{1.5cm}|}
 \caption{Estimated intelligence prediction performance (Pearson correlation/binary accuracy). Cell shading indicates the magnitude of the correlation.}
\label{table:performance_intelligence}\\
    \hline 
    Study &  intelligence & vocabulary & data & year & self-report\\
    \hline
    \endfirsthead
    \hline 
    Study &  intelligence & vocabulary & data & year & self-report\\
    \hline
    \endhead
\cite{RoblesGranda2020JointlyPJ} & \cellcolor[rgb]{0.9383263563333333,0.8089165520313726,0.741161515027451}.17/.55 & \cellcolor[rgb]
{0.9357737696666666,0.8122367012392158,0.7471564735843139}.16/.55 & private (multimodal) & 2020 & \checkmark\\
\hline
\cite{Titov2019FullscalePP} & \cellcolor[rgb]{0.9616447383764706,0.7580291825411765,0.6617823791647058}.27/.59 & \cellcolor[rgb]{0.968105337,0.6684751961176472,0.5504862209411766}.41/.63 & private (SN profile) & 2019 & \checkmark\\
\hgrayline
 & \cellcolor[rgb]{0.9692885689999999,0.6849817470823529,0.5689753262588235}.39/.63 & \cellcolor[rgb]{0.951253794882353,0.5787993895882353,0.4594078057647059}.53/.68 & private (SN subscriptions) &  & \checkmark\\
\hline
\cite{Wei2017HowSD} & \cellcolor[rgb]{0.9627082783294117,0.7535573465568628,0.655601211227451}.28/.59 & - & myPersonality (full) (\cite{Kosinski2013PrivateTA}) & 2017 & \checkmark\\
\hgrayline
\cite{Wei2017HowSD} & \cellcolor[rgb]{0.9698511524117647,0.6958300595294117,0.5813117740784314}.37/.62 & - & myPersonality (full) (\cite{Kosinski2013PrivateTA}) & 2017 &  $\times$\\
\hline
\end{longtable}

\captionsetup{width=20cm}
\begin{longtable}{|p{2.3cm}|p{0.9cm}|p{0.9cm}|p{0.9cm}|p{0.9cm}|p{0.9cm}|p{0.9cm}|p{0.9cm}|p{0.9cm}|p{0.9cm}|p{0.9cm}|p{2.6cm}|p{0.8cm}|p{.8cm}|}

\caption{Estimated prediction performance for portrait values\cite{schwartz2012overview}: achievement (A), benevolence (B), conformity (C), hedonism (H), power (P), security (S), self-direction (SD), stimulation (St), tradition (T), universalism (U) .  Estimates are reported as Pearson correlation/binary accuracy). Cell shading indicates the magnitude of the correlation.}
\label{table:performance_values}\\

\hline 
Study &  A & B & C & H & P & S & SD & St & T & U & data & year & self-report\\
\hline
\endfirsthead
\hline 
Study &  A & B & C & H & P & S & SD & St & T & U & data & year & self-report\\
\hline
\endhead

\cite{Titov2019FullscalePP} & \cellcolor[rgb]{0.9473454036,0.7946955048,0.7169905058}.20/.56 & \cellcolor[rgb]{0.9527607176705882,0.7829647976,0.6986457713058823}.22/.57 & \cellcolor[rgb]{0.9127650614705882,0.8366818943529412,0.7945121117647058}.10/.53 & \cellcolor[rgb]{0.9061541340352941,0.8420910651764706,0.8061505930823529}.08/.53 & \cellcolor[rgb]{0.9127650614705882,0.8366818943529412,0.7945121117647058}.10/.53 & \cellcolor[rgb]{0.925563423,0.8255172980705883,0.7711363078117647}.13/.54 & \cellcolor[rgb]{0.9193759889058823,0.8312727235294118,0.7828736304470588}.11/.54 & \cellcolor[rgb]{0.9688625003647059,0.7108384838117647,0.5999005044313725}.35/.61 & \cellcolor[rgb]{0.9226814526235294,0.8285681381176471,0.7770543897882353}.12/.54 & \cellcolor[rgb]{0.9193759889058823,0.8312727235294118,0.7828736304470588}.11/.54 & SN profile & 2019 & \checkmark\\
\hgrayline
 & \cellcolor[rgb]{0.9226814526235294,0.8285681381176471,0.7770543897882353}.12/.54 & \cellcolor[rgb]{0.9094595977529412,0.8393864797647058,0.8003313524235294}.09/.53 & \cellcolor[rgb]{0.9357737696666666,0.8122367012392158,0.7471564735843139}.16/.55 & \cellcolor[rgb]{0.9434315296666667,0.8022762536156862,0.7291715979137255}.18/.56 & \cellcolor[rgb]{0.9658988981882353,0.7401418386039216,0.6370577074156862}.30/.60 & \cellcolor[rgb]{0.9675442976352941,0.7308497161882352,0.6246854782352941}.32/.60 & \cellcolor[rgb]{0.9491505082901961,0.7907852690666667,0.7108755943019608}.21/.57 & \cellcolor[rgb]{0.9434315296666667,0.8022762536156862,0.7291715979137255}.18/.56 & \cellcolor[rgb]{0.9491505082901961,0.7907852690666667,0.7108755943019608}.21/.57 & \cellcolor[rgb]{0.9383263563333333,0.8089165520313726,0.741161515027451}.17/.55 & SN subscriptions & & \checkmark\\
 \hline
\cite{Kalimeri2019PredictingDM} & \cellcolor[rgb]{0.9473454036,0.7946955048,0.7169905058}.20/.56 & \cellcolor[rgb]{0.925563423,0.8255172980705883,0.7711363078117647}.13/.54 & \cellcolor[rgb]{0.9434315296666667,0.8022762536156862,0.7291715979137255}.18/.56 & \cellcolor[rgb]{0.9527607176705882,0.7829647976,0.6986457713058823}.22/.57 & \cellcolor[rgb]{0.9434315296666667,0.8022762536156862,0.7291715979137255}.18/.56 & \cellcolor[rgb]{0.9193759889058823,0.8312727235294118,0.7828736304470588}.11/.54 & \cellcolor[rgb]{0.9357737696666666,0.8122367012392158,0.7471564735843139}.16/.55 & \cellcolor[rgb]{0.9434315296666667,0.8022762536156862,0.7291715979137255}.18/.56 & \cellcolor[rgb]{0.9527607176705882,0.7829647976,0.6986457713058823}.22/.57 & \cellcolor[rgb]{0.9605811984235294,0.7625010185254902,0.6679635471019607}.26/.58 & mobile apps & 2019 & \checkmark\\
\hgrayline
& \cellcolor[rgb]{0.925563423,0.8255172980705883,0.7711363078117647}.13/.54 & \cellcolor[rgb]{0.9094595977529412,0.8393864797647058,0.8003313524235294}.09/.53 & \cellcolor[rgb]{0.9473454036,0.7946955048,0.7169905058}.20/.56 & \cellcolor[rgb]{0.9357737696666666,0.8122367012392158,0.7471564735843139}.16/.55 & \cellcolor[rgb]{0.925563423,0.8255172980705883,0.7711363078117647}.13/.54 & \cellcolor[rgb]{0.9434315296666667,0.8022762536156862,0.7291715979137255}.18/.56 & \cellcolor[rgb]{0.925563423,0.8255172980705883,0.7711363078117647}.13/.54 & \cellcolor[rgb]{0.925563423,0.8255172980705883,0.7711363078117647}.13/.54 & \cellcolor[rgb]{0.9434315296666667,0.8022762536156862,0.7291715979137255}.18/.56 & \cellcolor[rgb]{0.9473454036,0.7946955048,0.7169905058}.20/.56 & web domains &  & \checkmark\\
\hline
\end{longtable}

\captionsetup{width=20cm}
\begin{longtable}{|p{2.5cm}|p{5.4cm}|p{1.8cm}|p{1.5cm}|}
 \caption{Estimated prediction performance for various psychometric properties (Pearson correlation/binary accuracy). Cell shading indicates the magnitude of the correlation. }

\label{table:performance_other}\\

    \hline 
    Study &  Trait & performance & year\\
    \hline
    \endfirsthead
        \hline 
    Study &  Trait & performance & year\\
    \hline
    \endhead

\cite{Kalimeri2019PredictingDM} & Authority (Morals) & \cellcolor[rgb]{0.9688625003647059,0.7108384838117647,0.5999005044313725}.35/.61 & 2019\\
\hgrayline
 & Care (Morals) & \cellcolor[rgb]{0.9605811984235294,0.7625010185254902,0.6679635471019607}.26/.58 & \\
\hgrayline
 & Fairness (Morals) & \cellcolor[rgb]{0.9434315296666667,0.8022762536156862,0.7291715979137255}.18/.56 & \\
\hgrayline
 & Loyalty (Morals) & \cellcolor[rgb]{0.9648353582352941,0.7446136745882352,0.6432388753529412}.29/.59 & \\
\hgrayline
 & Purity (Morals) & \cellcolor[rgb]{0.9669624381411764,0.7356700026196078,0.6308765394784314}.31/.60 & \\
\hgrayline
 & Individualist/Binding (Morals) & \cellcolor[rgb]{0.9688625003647059,0.7108384838117647,0.5999005044313725}.35/.61 & \\
\hgrayline
 & Conservation (Values) & \cellcolor[rgb]{0.9648353582352941,0.7446136745882352,0.6432388753529412}.29/.59 & \\
\hgrayline
 & Openness (Values) & \cellcolor[rgb]{0.925563423,0.8255172980705883,0.7711363078117647}.13/.54 & \\
\hgrayline
 & Self-enhancement (Values) & \cellcolor[rgb]{0.9473454036,0.7946955048,0.7169905058}.20/.56 & \\
\hgrayline
 & Self-transcendence (Values) & \cellcolor[rgb]{0.9434315296666667,0.8022762536156862,0.7291715979137255}.18/.56 & \\
    \hline
\end{longtable}

\end{landscape}

\bibliography{mybib.bib}{}
\bibliographystyle{apa-good}

\end{document}